\documentclass[10pt,journal,compsoc]{IEEEtran}
\usepackage[pdftex]{graphicx}
\usepackage{booktabs, makecell, tabularx}
\usepackage{rotating}
\usepackage[export]{adjustbox}
\usepackage{gensymb}
\usepackage{ragged2e}
\usepackage{multirow}
\usepackage{xcolor}
\usepackage{color}
\definecolor{citecolor}{RGB}{0,0,0}
\usepackage[pagebackref=false,breaklinks=true,colorlinks,citecolor=citecolor,bookmarks=false]{hyperref}

\usepackage{soul}

\usepackage{cite}

\ifCLASSINFOpdf
  \usepackage[pdftex]{graphicx}
  \graphicspath{{../pdf/}{../jpeg/}{./figures}}
\else
\fi
\usepackage{xspace}
\usepackage{pgfplots}
\usepackage{xcolor}
\usepackage{tikz}

\usepackage{amsmath}
\usepackage{amssymb}
\usepackage{array}

\ifCLASSOPTIONcompsoc
  \usepackage[caption=false,font=footnotesize,labelfont=sf,textfont=sf]{subfig}
\else
  \usepackage[caption=false,font=footnotesize]{subfig}
\fi

\ifCLASSOPTIONcaptionsoff
  \usepackage[nomarkers]{endfloat}
 \let\MYoriglatexcaption\caption
 \renewcommand{\caption}[2][\relax]{\MYoriglatexcaption[#2]{#2}}
JOURNAL OF LATEX CLASS FILES, VOL. XX, NO. XX, October 2022
\fi
\usepackage{url}
\usepackage{etex}

\usepackage{pifont}
\usepackage[T1]{fontenc}

\begin{document}
\bstctlcite{IEEEexample:BSTcontrol}
%

\title{Not All Pixels Are Equal: Learning Pixel Hardness for Semantic Segmentation}

%
%
%
%

\author{Xin Xiao, Daiguo Zhou, Jiagao Hu, Yi Hu,Yongchao Xu
\IEEEcompsocitemizethanks{\IEEEcompsocthanksitem X. Xiao and Y. Xu are with National Engineering Research Center for Multimedia Software, Institute of Artificial Intelligence, School of Computer Science, and Hubei Key Laboratory of Multimedia and Network Communication Engineering, Wuhan University, Wuhan, 430070, China. (E-mail: \{xinxiao, yongchao.xu\}@whu.edu.cn).
\IEEEcompsocthanksitem D. Zhou and J. Hu are with Xiaomi AI Lab, Wuhan, 430070, China. (E-mail: \{zhoudaiguo, hujiagao\}@xiaomi.com).
\IEEEcompsocthanksitem Y. Hu is with Hubei Medical Devices Quality Supervision and Test Institute, Hubei Center for  AI Medical Devices Supercomputing, Wuhan, 430070, China. (E-mail: huyi@whmit.cn).}
\thanks{\textit{Corresponding author: Yongchao Xu}}
}

%
%

\markboth{Journal of \LaTeX\ Class Files,~Vol.~XX, No.~XX, October~2022}%
{Shell \MakeLowercase{\textit{et al.}}: Bare Demo of IEEEtran.cls for Computer Society Journals}
%



\newcommand{\ie}{\textit{i.e.}\xspace}
\newcommand{\eg}{\textit{e.g.}\xspace}
\newcommand{\etal}{\textit{et al.}\xspace}
\newcommand{\wrt}{\textit{w.r.t.}\xspace}
\newcommand{\etc}{\textit{etc.}\xspace}
\newcommand{\resp}{\textit{resp.}\xspace}

\newcommand{\fixedvskip}{-3mm}

\newcommand{\fixedvskiptab}{-2mm}

\IEEEtitleabstractindextext{%
\begin{abstract}
\justifying Semantic segmentation has recently witnessed great
  progress. Despite the impressive overall results, the segmentation
  performance in some hard areas (\eg, small objects or thin parts) is
  still not promising.  A straightforward solution is hard sample
  mining, which is widely used in object detection. Yet, most existing
  hard pixel mining strategies for semantic segmentation often rely
  on pixel's loss value, which tends to decrease during
  training. Intuitively, the pixel hardness for segmentation mainly
  depends on image structure and is expected to be stable.  In
  this paper, we propose to learn pixel hardness for semantic
  segmentation, leveraging hardness information contained in global
  and historical loss values. More precisely, we add a gradient-independent branch for learning a hardness level (HL) map by
  maximizing hardness-weighted segmentation loss, which is minimized
  for the segmentation head.  
  This encourages large hardness values in
  difficult areas, leading to appropriate and stable HL map.  Despite
  its simplicity, the proposed method can be applied to most
  segmentation methods with no and marginal extra cost during
  inference and training, respectively. Without bells
  and whistles, the proposed method achieves consistent/significant
  improvement (1.37\% mIoU on average) over most popular semantic
  segmentation methods on Cityscapes dataset, and demonstrates good
  generalization ability across domains.
  The source codes are available at
  \href{https://github.com/Menoly-xin/Hardness-Level-Learning}{https://github.com/Menoly-xin/Hardness-Level-Learning}.
\end{abstract}

}

\maketitle

\IEEEdisplaynontitleabstractindextext

%
\IEEEpeerreviewmaketitle

\IEEEraisesectionheading{\section{Introduction}\label{sec:introduction}}

\IEEEPARstart{S}{emantic} segmentation aims to assign a semantic label
to each pixel in an image, and is one of the fundamental tasks in
computer vision. Benefiting from some large-scale open-sourced
semantic segmentation datasets~\cite{cordts2016cityscapes,
  zhou2019semantic, waqas2019isaid} and developments of backbone
networks~\cite{simonyan2014very, he2016deep, tan2019efficientnet,
  liu2021swin}, numerous studies have made impressive progress in the
field of semantic segmentation. Specifically, since
FCN~\cite{long2015fully} shed new light on pixel-wise prediction in an
end-to-end manner using a fully convolutional neural network, enormous
efforts have been devoted to developing new dense prediction style
segmentation architectures. For example, PSPNet~\cite{zhao2017pyramid}
fuses multi-scale feature maps for more sophisticated feature
representation. The DeepLab family~\cite{chen2017deeplab,
  chen2017rethinking, chen2018encoder} enlarges receptive field via
astrous spatial pyramid pooling (ASPP). Recently, many
studies~\cite{wang2018non, li2018pyramid, fu2019dual,
  li2019expectation, he2019dynamic, zhu2019asymmetric, cao2019gcnet,
  huang2019interlaced, zhong2020squeeze, huang2020ccnet,
  yin2020disentangled, li2021ctnet, liu2022covariance} resort to the
attention mechanism for gathering more context information from the
whole image for better semantic segmentation.

\begin{figure*}[htp]
  \centering
  \includegraphics[width=\linewidth]{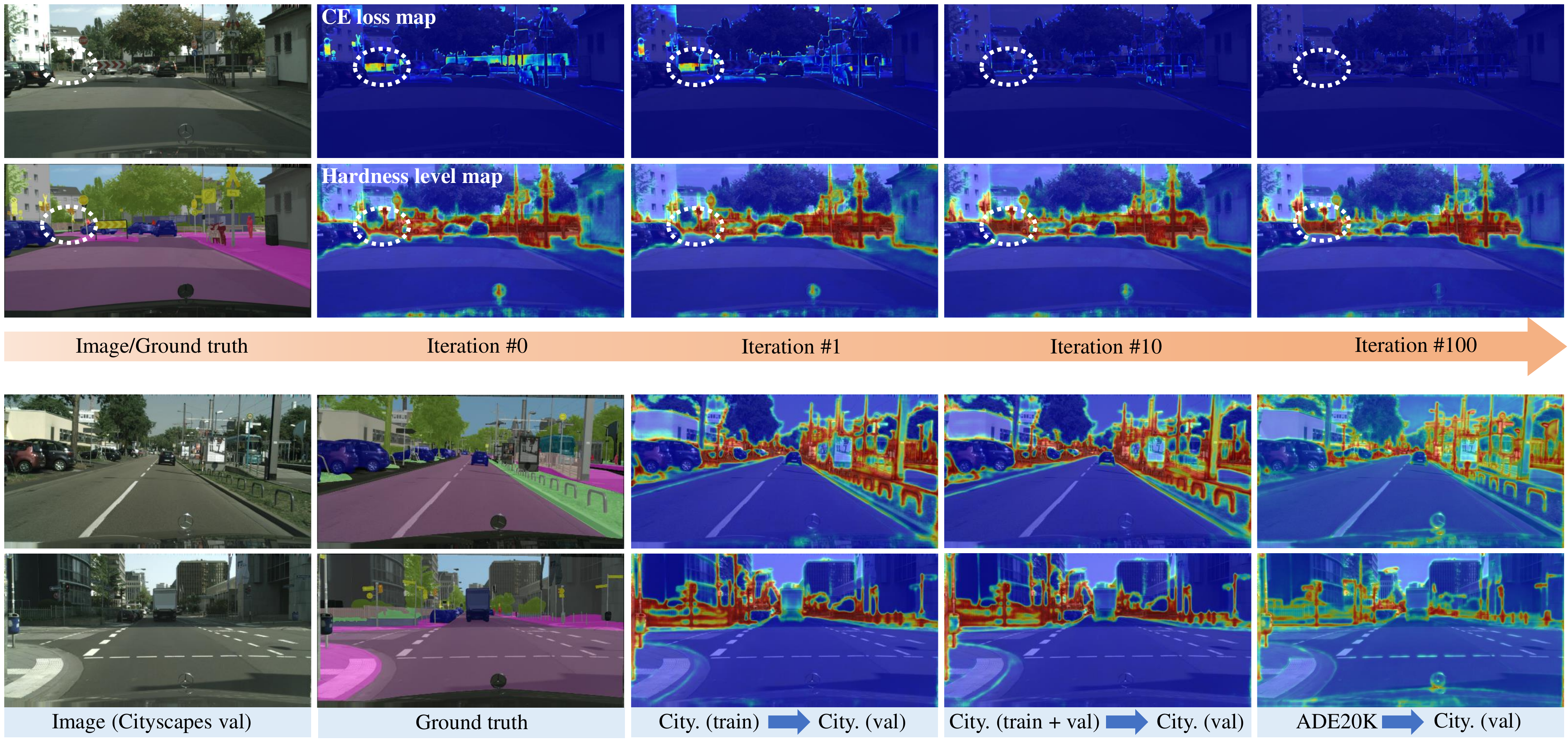}
  \vspace{-5mm}
  \caption{Visualization of cross-entropy loss maps and learned
    hardness level (HL) maps at different iterations (top two rows) of
    continuing to train the model (well trained on Cityscapes training
    set) on an image in the validation set of Cityscapes.
    The bottom two rows show some learned HL maps on images in
    the validation set of Cityscapes using three models trained on
    Cityscapes training set, Cityscapes training and validation set,
    and ADE20K training set, respectively. The learned HL maps are
    related to inherent image structure, and are thus quite and
    generalize well.}
  \label{fig:hardness}
\end{figure*}

These methods effectively boost the performance of semantic
segmentation by a large margin. Yet, most of them mainly frame the
segmentation task as individual pixel-wise classification tasks,
calculating the loss value for each pixel and then equally-weighted
averaging the loss values to get an image-level loss. Such a scheme
ignores that the difficulty in classifying various pixels in an image
is quite different. In fact, semantic segmentation is with a structured
output. Many pixels are relatively easy to segment. The area with
a complex structure deserves more attention for both manual annotation
and segmentation. Therefore, a straightforward idea is to apply larger
weights to harder pixels in averaging the pixel-wise loss values.

Over the past several years, the topic of focusing more on hard
samples during training has attracted great research
interest~\cite{shrivastava2016training, lin2017focal,
  li2022generalized} in object detection. Existing
methods~\cite{shrivastava2016training, lin2017focal,
  li2022generalized} typically rely on current loss values to
characterize the hardness of different samples, and only make use of
hard samples with large loss values for
training~\cite{shrivastava2016training} or assign larger loss weights
to samples with larger loss values~\cite{lin2017focal,
  li2022generalized}. These hard sample mining methods effectively
address the problem of extremely unbalanced hard and easy samples,
achieving impressive results in object detection.

Different from object detection, to the best of our knowledge, 
hard pixel mining is not widely used in semantic
segmentation. Directly applying hard pixel mining strategies based
on current pixel loss values~\cite{shrivastava2016training,
  lin2017focal, li2022generalized} to semantic segmentation may even
harm the segmentation performance. In fact, thanks to the strong
memorization ability of deep neural networks~\cite{arpit2017closer},
the segmentation network is capable to well fit most pixels during
training. Only a very few pixels around object boundaries have large
loss values (see the top row of Fig.~\ref{fig:hardness}). Yet, it is
usually ambiguous to accurately distinguish these pixels around
boundaries even for manual annotation. The up-sampling operation in
segmentation network further makes it ambiguous to discriminate these
pixels for segmentation. The training focused on these ambiguous
pixels may lead to degenerated segmentation results. Therefore,
trivially paying more attention to pixels with large loss values is
not very effective.

There exist some semantic segmentation methods~\cite{li2017not,
  wang2020hard, yin2019online, deng2022nightlab, chu2020pay,
  nie2020adversarial, hu2019topology, hu2021topology} that consider
the difference of segmentation difficulty for different pixels in a
more sophisticated way. Specifically, some methods~\cite{li2017not,
  wang2020hard} regard regions with confident (having  large maximum
scores of predicted probability) segmentation results as easy regions
and unconfident regions as hard regions, and then apply different
segmentation heads for hard and easy regions. Some other methods
select hard regions based on loss values~\cite{yin2019online} or an
object detection network~\cite{deng2022nightlab}, and then zoom in the
selected hard regions for refined segmentation. In medical image
segmentation, some methods adopt an extra segmentation loss on
non-discontinuity areas~\cite{chu2020pay} or regions with anatomically
implausible segmentation results based on adversarial confidence
learning~\cite{nie2020adversarial} or topological
analysis~\cite{hu2019topology, hu2021topology}. Taking into account
the difference of segmentation difficulty among
pixels~\cite{li2017not, wang2020hard, yin2019online, deng2022nightlab,
  chu2020pay, nie2020adversarial, hu2019topology, hu2021topology} have
been proven useful in improving segmentation results.

In this paper, different from the existing methods, we propose to
learn pixel hardness for semantic segmentation. More precisely,
instead of relying on the current loss values for extracting hard
regions, we propose to duplicate the segmentation head for learning a
hardness level (HL) map, based on which we minimize a
hardness-weighted segmentation loss for the segmentation head. Since
there is no ground-truth for the pixel hardness, the key lies on how
to learn an appropriate hardness level map. For that, we control the
maximum relative ratio between the largest and smallest hardness level
among all pixels by applying a Sigmoid operation and adding a small
constant value. Besides, we also normalize the sum of harness level
over all pixels to 1. This also acts as a hardness level competition
among pixels, avoiding only a few overwhelming pixels with large
hardness level. We then maximize the hardness-weighted segmentation
loss for optimizing the HL branch. Since deep neural networks usually
quickly fit easy samples and gradually fit hard
samples~\cite{chatterjee2022generalization}, maximizing the
hardness-weighted segmentation encourages larger hardness level on
pixels with larger historical loss values. As illustrated in the
bottom two rows of Fig.~\ref{fig:hardness}, this leads to a stable and
meaningful hardness level map, which is related to the image structure
and thus generalizes well to unseen images. Paying more attention to
pixels with larger hardness level results in improved segmentation
results. Note that the HL branch is only involved in training phase
and has independent gradient flow with the segmentation head.

The main contribution of this paper is three-fold: 
1) We propose a
novel idea of learning pixel hardness for semantic segmentation,
subtly making use of the difference of difficulty in segmenting
different pixels; 2) We develop an effective scheme in learning the
hardness level map without explicit ground-truth, yielding stable and
meaningful hardness level map which generalizes well to unseen images;
3) Without bells and whistles, the proposed method achieves
consistent/significant improvements over most semantic segmentation
methods on various types of images, and demonstrates great potential
in semi-supervised and domain-generalized semantic segmentation.


\section{Related Work}
\label{sec:related}

\subsection{Semantic segmentation}
\label{subsec:segmentation}
Semantic segmentation has been studied for a long
time~\cite{minaee2021image}. The pioneering work
FCN~\cite{long2015fully} firstly introduces fully convolutional
networks and renders the segmentation task end-to-end manner, opening
up new avenues for semantic segmentation. Since then, numerous efforts
have been devoted to developing convolutional segmentation
methods. For instance, some early studies~\cite{chen2017deeplab,
  zheng2015conditional} adopt structured operators (\eg, conditional
random fields) to refine segmentation results, but at the price of a
substantial increase in inference cost. Some following
works~\cite{zhao2017pyramid, chen2017rethinking, chen2018encoder,
  xiao2018unified, YuanCW20, liu2021holistically, yu2021bisenet,
  yuan2021ocnet} design advanced segmentation architectures for better
feature representation. For example, PSPNet~\cite{zhao2017pyramid}
fuses multi-scale features by pyramid pooling module. DeepLab
series~\cite{chen2017rethinking, chen2018encoder} adopts atrous
spatial pyramid pooling to enlarge receptive field. Since the
introduction of attention mechanism to the vision
task~\cite{wang2018non}, numerous studies~\cite{fu2019dual,
  zhong2020squeeze, li2018pyramid, wang2018non, li2019expectation,
  huang2020ccnet, he2019dynamic, zhu2019asymmetric, cao2019gcnet,
  huang2019interlaced, yin2020disentangled, li2021ctnet,
  liu2022covariance} have focused on this topic. Most of them mainly
employ the non-local operator or the attention mechanism for gathering
semantic context information across the whole image. Some
methods~\cite{yu2018learning, borse2021inverseform, wang2021active}
attempt to improve the segmentation accuracy by better aligning
boundary or leverage boundary-related features. Recently,
transformer-based semantic segmentation
methods~\cite{zheng2021rethinking, xie2021segformer} have also
attracted much attention and achieved impressive segmentation results.

With the impressive development of segmentation architectures,
backbone networks also evolve rapidly. Since
AlexNet~\cite{krizhevsky2012imagenet} brings computer vision into a
new era, convolutional neural networks serve as commonly used
backbones throughout recent computer vision tasks. Many convolutional
backbone networks (\eg,VGG~\cite{simonyan2014very},
ResNet~\cite{he2016deep}, Res2Net~\cite{gao2019res2net},
HRNet\cite{wang2021deep}) being deeper and more effective have been
proposed. Besides, compact and efficient backbone networks (\eg,
MobileNet~\cite{sandler2018mobilenetv2},
EfficientNet~\cite{tan2019efficientnet}) which require much less
running cost have also been proposed, making it possible to deploy
convolutional neural networks on low-performance equipment. More
recently, some vision
transformers~\cite{dosovitskiy2020image,liu2021swin,bao2021beit}
greatly boost segmentation performance. The CNN-based network
ConvNeXt~\cite{liu2022convnet} with modern components also achieves
comparable performance with some vision transformers.

Benefiting from these advanced studies, semantic segmentation has
witnessed considerable progress. Yet, most studies equally consider
each pixel in segmentation, ignoring that the difficulty for
segmenting each pixel in an image is quite different. The proposed
method learns the pixel hardness for semantic segmentation,
effectively making use of the difference of segmentation difficulty
among pixels. The proposed method can be plugged to most semantic
segmentation methods, improving the performance with ignorable extra
training cost and no extra cost during test.

\subsection{Hard sample mining}
\label{subsec:mining}

Some early studies~\cite{shrivastava2016training, lin2017focal,
  li2022generalized} have noticed that the difficulty of classifying
different samples in an image is quite different in the field of
object detection. They adjust the weights for different samples based
on their loss values. For example, OHEM~\cite{shrivastava2016training}
only picks out hard samples with high loss values for training, which
can be viewed as assigning 0 weight for the other easy samples with
small loss values. Focal loss~\cite{lin2017focal} assigns much larger
weights for hard samples with higher loss values, and much smaller
weights for easy samples with lower loss values. These hard mining
strategies have made impressive progress in object detection. Yet, to
the best of our knowledge, there does not exist a widely used
loss-value-based weighting method specifically designed for semantic
segmentation.

In the field of semantic segmentation, there are also some
works~\cite{li2017not, wang2020hard, yin2019online, deng2022nightlab,
  chu2020pay, nie2020adversarial, hu2019topology, hu2021topology}
follow the spirit of hard sample mining. For instance, Li
\etal~\cite{li2017not} present a deep layer cascade method that adopts
shallow (\resp deep) network for easy (\resp hard) regions with large
(\resp small) maximum scores of predicted probability. The work
in~\cite{wang2020hard} applies three segmentation heads for coarse
segmentation, segmentation on hard and easy regions also based on the
maximum score of predicted probability. The segmentation results of
the three heads are then fused together to generate the final
segmentation. OHRM~\cite{yin2019online} and
NightLab~\cite{deng2022nightlab} first extract hard regions based on
loss values~\cite{yin2019online} or an detection
network~\cite{deng2022nightlab}, then zoom in the extracted hard
regions for re-training~\cite{yin2019online} or segmentation
refinement in both training and testing~\cite{deng2022nightlab}.

Some medical image segmentation works~\cite{nie2020adversarial,
  chu2020pay, hu2019topology, hu2021topology} perform hard pixel mining
by assigning larger loss weights to harder areas. For instance, Nie
and Shen~\cite{nie2020adversarial} rely on adversarial confidence
learning by using a discriminator on the segmentation output to find hard
regions from the aspect of shape structure, and then apply a
difficulty-aware attention mechanism on the hard
regions. In~\cite{chu2020pay}, the authors propose to simply add extra
loss in non-discontinuity areas. The studies in~\cite{hu2019topology,
  hu2021topology} attempt to locate topologically important areas via
topological analysis of the segmentation output, and apply extra loss
on these areas. These works in~\cite{chu2020pay, hu2019topology,
  hu2021topology} are equivalent to assigning larger loss weights on hard
regions given by non-discontinuity or topologically important areas.

The works in~\cite{shrivastava2016training, lin2017focal,
  nie2020adversarial, chu2020pay, hu2019topology, hu2021topology} are
the most related works for the proposed method. Different
from~\cite{shrivastava2016training, lin2017focal} that characterize
the hardness based on current loss values, the proposed method
automatically learns the pixel hardness and makes use of historical
and global loss values, leading to more appropriate loss weights and
better segmentation performance. Compared
with~\cite{nie2020adversarial, chu2020pay, hu2019topology,
  hu2021topology} which aim to segment medical objects often having
specific prior shapes, the proposed method is more general and able to
segment objects in natural images whose shape structure varies much
more than medical objects.

\begin{figure*}[htp]
  \centering
  \includegraphics[width=\linewidth]{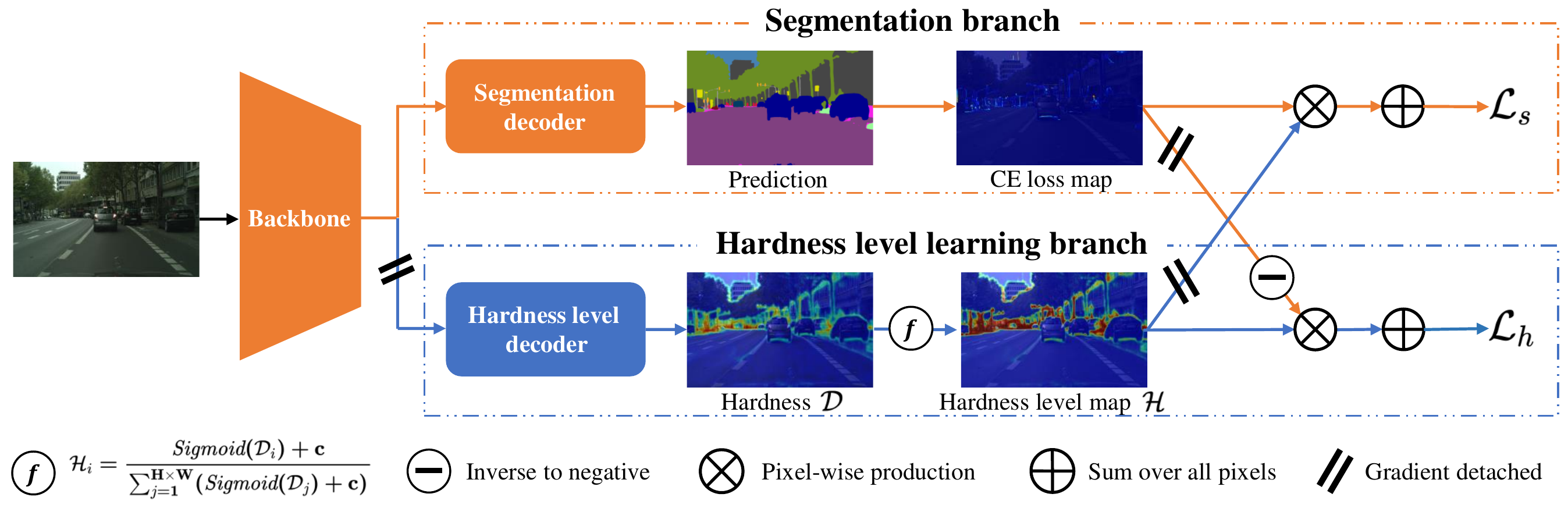}
  \caption{The pipeline of the proposed pixel hardness learning
    method. The hardness level learning branch shares the same network
    architecture as the segmentation head, and only works during
    training phase and thus requires no cost during testing
    phase. Note that the gradient flow separately propagates along the
    corresponding branch (the path of the same color), avoiding the
    gradient interaction between the two branches.}
  \label{fig:overview}
\end{figure*}

\section{Proposed method}
\label{sec:method}

\subsection{Overview}
\label{subsec:overview}
Existing semantic segmentation methods mainly adopt equally-weighted
average loss for all pixels in an image to get the image-level
loss. This ignores that the difficulty in segmenting each pixel in an
image is different, which also holds for manual annotation. Based on
this, a straightforward idea is to apply different loss weights for
pixels of different hardness. Indeed, the hard sample mining strategy
which assigns higher loss weights for samples with higher loss values,
has been proven very useful in object detection~\cite{lin2017focal,
  shrivastava2016training, li2022generalized}.
Yet, to the best of our knowledge, the effective hard sample mining
in object detection is not widely used in semantic
segmentation. Directly adopting such hard pixel mining usually yields
degenerated segmentation results. In fact, deep neural networks have extraordinary memorization
ability~\cite{arpit2017closer}, and are capable of memorizing almost all
training samples. This results in very few pixels with large loss
values during the segmentation training. Most of these pixels
with large loss values lie around object boundary (see
Fig.~\ref{fig:hardness}), where it is often ambiguous to distinguish
different classes for both annotation and segmentation due to
inevitable up-sampling operation. The optimization focused mainly on
these few pixels may lead to a wrongly over-fitted segmentation model.
Therefore, the hardness based on current loss value is not effective
for semantic segmentation.

To better take into account the difference of segmentation difficulty,
we propose to learn pixel hardness for semantic
segmentation. Specifically, we propose a novel hardness level learning
method to extract pixel hardness knowledge from the evolving
historical pixel loss values. In fact, deep neural networks usually
start to quickly fit easy samples and then gradually fit hard
samples~\cite{chatterjee2022generalization}. This fitting process
contains rich information about pixel hardness. Based on this, we introduce an auxiliary hardness level (HL) learning
branch, accumulating historical information.
This branch is optimized by maximizing the hardness-weighted segmentation loss given by
the multiplication of hardness level map and cross-entropy loss map. Such an optimization scheme
encourages to assign high hardness values for pixels with large loss
values in the training process, and makes use of the global and
historical pixel loss values. This results in a stable and meaningful
hardness level map related to the inherent structure of an image (see
Fig.~\ref{fig:hardness}). For the segmentation branch, we minimize the
hardness-weighted segmentation loss. Note that the segmentation branch and HL
learning branch have independent gradient flow during optimization in
training. The HL learning branch is only involved during training, and
thus requires no extra cost in inference. The overall pipeline is
depicted in Fig.~\ref{fig:overview}.

\subsection{Pixel hardness learning}
\label{subsec:hardnesslearning}
The mainstream semantic segmentation methods mainly adopt a backbone
convolutional neural network to extract multi-scale features, followed
by a segmentation head composed of several convolution layers. The
segmentation usually relies on a $1 \times 1$ convolution layer for
pixel-wise classification, using equally averaged cross-entropy loss
over all pixels. This ignores that the segmentation difficulty for
different pixels is not equal for both manual annotation and
segmentation. Indeed, during annotation, we usually pay more
attention to complex areas. The segmentation network first quickly
fits easy pixels, and then gradually fits hard
pixels~\cite{chatterjee2022generalization}. We propose to make use of
this property to learn pixel hardness for semantic segmentation. The
key lies on how to learn the hardness level (HL) map without direct
and explicit ground-truth supervision. We detail the network
architecture and training objective in the following.

\medskip
\noindent
\textbf{Network architecture for HL learning:} In addition to the
segmentation head, we introduce an auxiliary hardness level learning
branch on the extracted multi-scale feature of the backbone network.
For the sake of simplicity, we simply adopt the same network
architecture as the segmentation head by changing the output channel
to 1 for learning pixel hardness $\mathcal{D}$. As depicted in
Fig.~\ref{fig:overview}, we apply a transformation on the hardness
$\mathcal{D}$ to obtain the final hardness level map. Specifically, we
apply a Sigmoid function on $\mathcal{D}$ to make it into range [0,
  1]. This avoids negative hardness and too large overwhelming
hardness. Besides, we also add a constant weight $c$ to the output of
Sigmoid function, avoiding some pixels being neglected for too small
$\mathcal{D}$. In fact, the constant $c$ in $\mathcal{Z} = Sigmoid
(\mathcal{D}) + c$ acts as a hyper-parameter that sets lower/upper
bound for pixel hardness level, and controls the maximal relative
hardness ratio between different pixels. This further prevents
overwhelming hardness for some pixels. We then divide $\mathcal{Z}$ by
the sum of $\mathcal{Z}$ over all pixels in the image, yielding the
final hardness level map $\mathcal{H}$. Formally, for the $i$-th
pixel, the final hardness level $\mathcal{H}_i$ is given by:
\begin{equation}
\begin{split}
\label{eq:hardnesslevel}
  \mathcal{H}_{i} \, = \, \frac{\mathcal{Z}_{i}}{\sum_{j = 1}^{H \times W} \mathcal{Z}_j} = \frac{Sigmoid(\mathcal{D}_i) + c}{\sum_{j = 1}^{H \times W}\big(Sigmoid(\mathcal{D}_j) + c\big)},
\end{split}
\end{equation}
where $H$ and $W$ stand for the height and width of the image,
respectively. This normalization makes the sum of the hardness level
over all pixels equal 1, acting also as a competition mechanism for
hardness levels among all pixels in the image. A high hardness level
for a pixel relatively limits the hardness level for the other pixels.

Note that the auxiliary hardness level learning branch is detached
(see Fig.~\ref{fig:overview}). In this way, the extra branch does not
influence the shared backbone for segmentation branch.

\medskip
\noindent
\textbf{Training objective for HL learning:} Since there is no
ground-truth for the hardness level, the key of HL learning lies on
how to design an appropriate training objective. Considering that deep
neural networks usually begin to quickly fit easy samples and then
gradually fit hard samples during training process, the pixel-wise
cross-entropy loss values of classical segmentation loss $\mathcal{L}$
quickly converge to small values for easy pixels, and keep relatively
large in many training iterations for hard pixels (see the first row
in Fig.~\ref{fig:hardness}). Therefore, the historical pixel-wise
cross-entropy loss values $\mathcal{L}$ during training process
encodes the pixel hardness information. Based on this, we propose to
minimize the following loss for the HL learning branch:
\begin{equation}
\begin{split}
\label{eq:lossh}
  \mathcal{L}_{\textit{h}} \, = \, -\sum_{i=1}^{H \times W} \mathcal{H}_{i}\times \mathcal{L}^{d}_{i},
\end{split}
\end{equation}
where $\mathcal{L}^d$ denotes the gradient detached cross-entropy loss
$\mathcal{L}$. As depicted in Fig.~\ref{fig:overview}, the gradient
flow of minimizing $\mathcal{L}_h$ does not back-propagate to the
segmentation branch, and only propagates along the auxiliary HL
learning branch.

Minimizing the loss function in Eq.~\eqref{eq:lossh} is equivalent to
maximize the hardness-weighted cross-entropy segmentation loss, which
encourages high hardness level for hard pixels with large historical
loss values. Specifically, the gradient of the loss function defined
in Eq.~\eqref{eq:lossh} with respect to the hardness level is given
by:
\begin{equation}
  \begin{split}
    \label{eq:gradient}
    \frac{\partial \mathcal{L}_h}{\partial \mathcal{H}} \, = \, -\mathcal{L}^d.
  \end{split}
\end{equation}
From the aspect of gradient, since the cross-entropy loss value on
each pixel is non-negative, minimizing the loss $\mathcal{L}_h$ mainly
leads to a competitive increase of hardness level for all pixels, on
which the summation of hardness level equals 1 based on
Eq.~\eqref{eq:hardnesslevel}. A higher (\resp very small) pixel loss
value triggers relatively larger (\resp ignorable) increasing of the
hardness level on the corresponding pixel. Therefore, minimizing the
loss function $\mathcal{L}_h$ results in large (\resp small) hardness
level for hard (\resp easy) pixels with relatively high (\resp low)
historical pixel loss values. Besides, thanks to the Sigmoid operation
and the hyper-parameter $c$ in Eq.~\eqref{eq:hardnesslevel} that avoid
overwhelming hardness for some pixels, the set of pixels with
relatively large historical loss values would have high hardness
level. This gives rise to relatively stable and meaningful hardness
level related to the inherent structure of an image (see the second row
of Fig.~\ref{fig:hardness}).

\subsection{Hardness-aware semantic segmentation}
\label{subsec:hardnessseg}
The proposed pixel hardness learning can be applied to most popular
semantic segmentation methods. We keep the segmentation head unchanged.
Instead of classical equally-weighted cross-entropy loss used in most
semantic segmentation methods, we propose to minimize the following
hardness-weighted segmentation loss $\mathcal{L}_s$:
\begin{equation}
\begin{split}
\label{eq:lossseg}
  \mathcal{L}_s \, = \, \sum_{i=1}^{H \times W} \mathcal{H}^{d}_{i} \times \mathcal{L}_i,
\end{split}
\end{equation}
where $\mathcal{H}^d$ stands for the gradient detached hardness level
map. As shown in Fig.~\ref{fig:overview}, the gradient flow of
minimizing $\mathcal{L}_s$ only back-propagates along the segmentation
branch, and does not influence the hardness level learning branch.

The final overall loss $\mathcal{L}_f$ for the whole model is given by.
\begin{equation}
\begin{split}
\label{eq:lossfinal}
  \mathcal{L}_{f} \, = \, \mathcal{L}_s + \alpha \times \mathcal{L}_h,
\end{split}
\end{equation}
where $\alpha$ is a hyper-parameter that scales the learning rate for
the hardness level learning branch. Based on Eq.~\eqref{eq:lossh} and
Eq.~\eqref{eq:lossseg}, the hardness level learning and segmentation
branch are separately optimized. The $\alpha$ controls the increasing
rate of competing hardness level on each pixel during training.

\section{Experiments}
\label{sec:exp}

\subsection{Datasets and evaluation protocol}
\label{subsec:dataset}
We conduct extensive experiments on
Cityscapes~\cite{cordts2016cityscapes} and
ADE20K~\cite{zhou2019semantic} for natural image segmentation,
{iSAID}~\cite{waqas2019isaid} and {Total-Text}~\cite{CK2019} for
aerial and text image segmentation, respectively. The details of these
datasets are given as follows.

\medskip
\noindent\textbf{Cityscapes}~\cite{cordts2016cityscapes} is a high
quality semantic segmentation dataset for urban scene understanding,
which contains 5,000 finely annotated images (2,975, 500, and 1,525
for training, validation, and test set, respectively) and about 20,000
coarsely annotated images. Only the finely annotated are used in
training for all experiments. We mainly report segmentation results on
the validation set.

\medskip
\noindent\textbf{ADE20K}~\cite{zhou2019semantic} is a widely used
scene parsing benchmark dataset containing pixel-wise annotations of
150 categories. This dataset is pretty challenging due to its numerous
classes. The dataset provides 20,000 and 2,000 images for training and
validation, respectively.

\medskip
\noindent\textbf{iSAID}~\cite{waqas2019isaid} is a large-scale aerial
image dataset for semantic segmentation. This dataset contains 2,806
high-resolution images with segmentation annotations of 15
categories. The dataset is split into 1/2, 1/6, and 1/3 portion for
training, validation, and test, respectively. Following the common
practice, we cut the original high-resolution images into small
patches, and report results on the validation set.

\medskip

\noindent\textbf{Total-Text}~\cite{CK2019} consists of 1,555 images
with more than 3 different text orientations, including horizontal,
multi-oriented, and curved. There are 1,255 and 300 images for
training and testing, respectively. The annotations only contain two
categories: foreground texts and background.

\medskip
\noindent
\textbf{Evaluation protocol: } We adopt the classical mean of
class-wise intersection over union (mIoU) for all quantitative
evaluation of segmentation performance.

\begin{table}[tp]
  \centering
  \caption{Training details and test strategy on different
    datasets. We adopt the same settings for the baseline methods and
    the proposed approach.}
  {
  \begin{tabular*}{\linewidth}{lclcc}
    \toprule
    \multirow{2}*{Dataset}  &\multicolumn{3}{c}{Training details}& \multirow{2}*{Test strategy} \\
    \cmidrule{2-4}
       &Batch size &\#Iterations &Crop size &  \\
    \midrule
    Cityscapes~\cite{cordts2016cityscapes} &8 &40,000 & $769\times 769$ &Sliding\\
    \midrule
    ADE20K~\cite{zhou2019semantic} &16 &160,000 & $512\times 512$ &Whole\\
    \midrule
    iSAID~\cite{waqas2019isaid} &16 &80,000 & $896\times 896$ &Whole\\
    \midrule
    Total-Text~\cite{CK2019} &16 &40,000 & $512\times 512$ &Whole\\
    \bottomrule
  \end{tabular*}
  }
  \label{tab:implement_detail}
\end{table}

\subsection{Implementation details}
\label{subsec:implementation}
The proposed method is implemented using the
mmsegmentation~\cite{mmseg2020} framework on a workstation with 8
NVIDIA Tesla A100 GPUs. We adopt the default settings (including batch
size, number of iterations, crop size, test strategy, optimizer and
related parameters for optimization, etc.) of mmsegmentation for all
baseline methods and the proposed method. Unless explicitly stated, we
follow the most common default settings (listed in
Tab.~\ref{tab:implement_detail}) of mmsegmentation.

During training, we augment the training images with common data
augmentation strategies, including random scaling between 0.5 and 2.0,
random horizontal flipping, random cropping, and random color
jittering.

In all experiments, the hyper-parameter $c$ involved in
Eq.~\eqref{eq:hardnesslevel} is set to 0.1. We set the hyper-parameter
$\alpha$ involved in Eq.~\eqref{eq:lossfinal} to 0.01 for all
experiments. Note that we report the semantic segmentation performance
of \textit{single scale} inference using the model of the last iteration
for all baseline methods and the proposed method.

\subsection{Analysis of learned hardness level map}
\label{subsec:validity}

We conduct three types of analysis on the learned HL maps. In the
following, we first visualize the learned HL maps, followed by
an analysis of the relation between HL maps and segmentation quality. We
then show the generalization ability of HL map to unseen images.

\begin{figure}[tp]
\centering
\subfloat[Image]{
	\begin{minipage}[b]{0.32\linewidth}
		\includegraphics[width=1\linewidth]{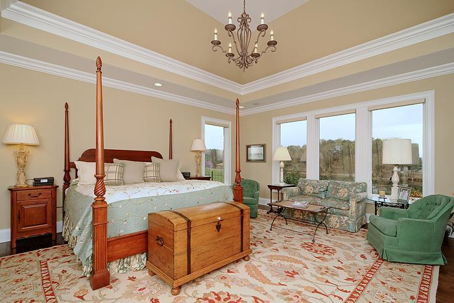}\vspace{0.75mm}
		\includegraphics[width=1\linewidth]{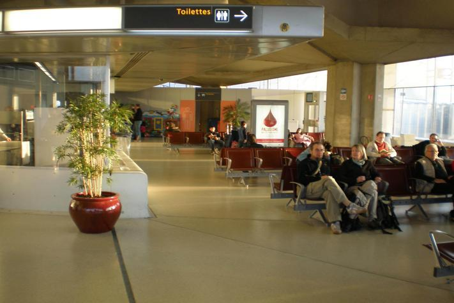}\vspace{0.75mm}
            \includegraphics[width=1\linewidth]{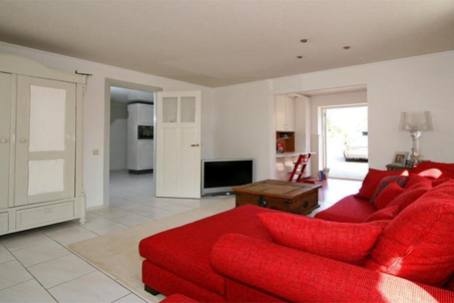}\vspace{0.75mm}
		\includegraphics[width=1\linewidth]{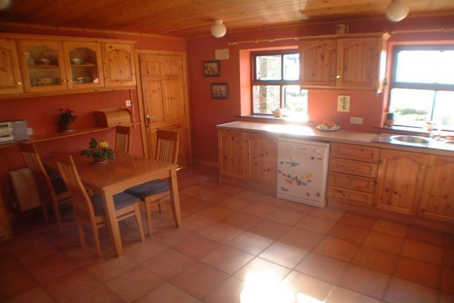}\vspace{0.75mm}
	\end{minipage}
} 
\hspace{-2mm}
\subfloat[Ground truth]{
	\begin{minipage}[b]{0.32\linewidth}
		\includegraphics[width=1\linewidth]{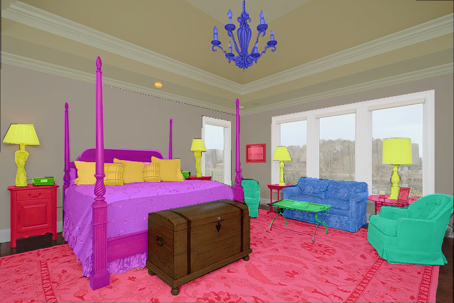}\vspace{0.75mm}
		\includegraphics[width=1\linewidth]{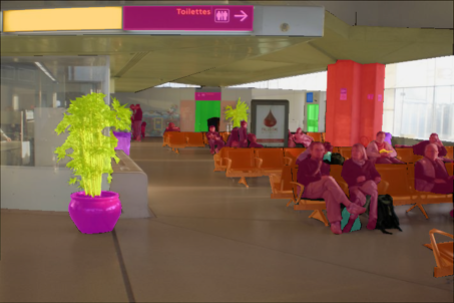}\vspace{0.75mm}
            \includegraphics[width=1\linewidth]{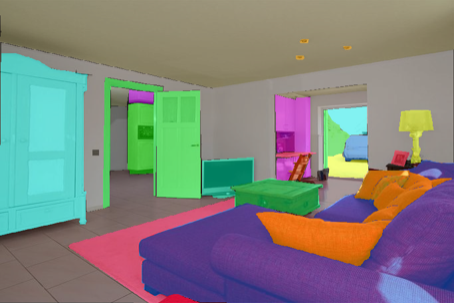}\vspace{0.75mm}
		\includegraphics[width=1\linewidth]{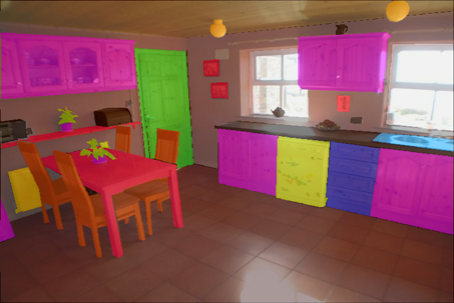}\vspace{0.75mm}
	\end{minipage}
} 
\hspace{-2mm}
\subfloat[HL map]{
	\begin{minipage}[b]{0.32\linewidth}
		\includegraphics[width=1\linewidth]{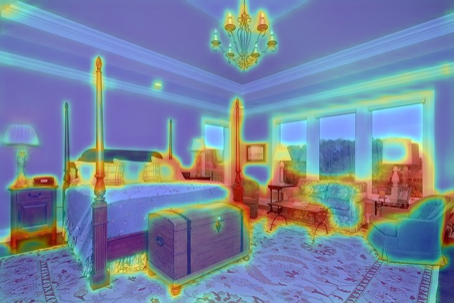}\vspace{0.75mm}
		\includegraphics[width=1\linewidth]{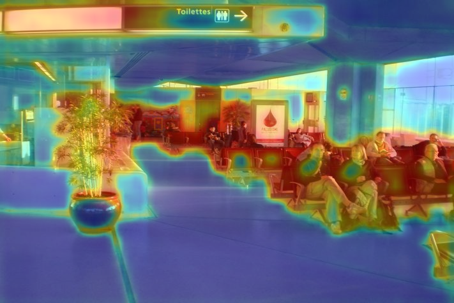}\vspace{0.75mm}
            \includegraphics[width=1\linewidth]{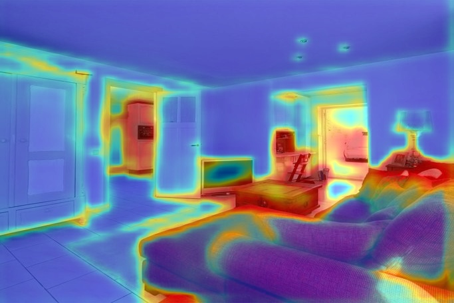}\vspace{0.75mm}
		\includegraphics[width=1\linewidth]{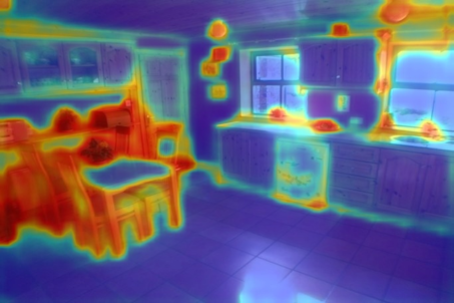}\vspace{0.75mm}
	\end{minipage}
} 
  \caption{Visualization of some learned hardness level maps for images in training (top two rows) and  validation (bottom two rows) set of ADE20K.}
  \label{fig:hardness_ade}
\end{figure}

\medskip
\noindent\textbf{Visualization of meaningful and stable HL maps: } The
proposed method aims to learn pixel hardness for semantic
segmentation. We first visualize some learned HL maps for images in
the training set of a model. As shown in Fig.~\ref{fig:hardness}
(second column from the right in bottom two rows) and
Fig.~\ref{fig:hardness_ade} (top two rows), the learned HL maps have
large values on complex areas, where they are difficult for segmentation
and more attention should be paid for manual annotation. Therefore,
the learned HL map is somehow meaningful and related to the inherent
structure of an underlying image.

We also visualize how the HL map is learned during training. For that,
we conduct an experiment by continuing to train a model (well-trained
on Cityscapes training set) on an image in the validation set of
Cityscapes for 100 iterations. As illustrated in
Fig.~\ref{fig:hardness} (top two rows), the cross-entropy loss values
tend to decrease to very small values. The HL map is in general rather
stable. Taking a closer look on the region within the white ellipse in
Fig.~\ref{fig:hardness}, the HL values increase on pixels with
relatively large historical pixel-wise cross-entropy loss
values. Starting from the 10-th iteration in Fig.~\ref{fig:hardness},
almost all pixels have very low cross-entropy loss values. The learned
HL map stays very stable from the 10-th iteration to the 100-th
iteration.  As explained in Sec.~\ref{subsec:hardnesslearning}, such
behavior of the HL map is reasonable based on Eq.~\eqref{eq:gradient}
and Eq.~\eqref{eq:hardnesslevel}. This implies that the proposed HL
learning method effectively makes use of the historical loss values
rather than current loss values, yielding a stable and meaningful HL map
related to image structure.

\begin{figure}[!tbp]
  \centering
    \subfloat[Cityscapes]{
	\begin{minipage}[b]{0.47\linewidth}
		\includegraphics[width=\linewidth]{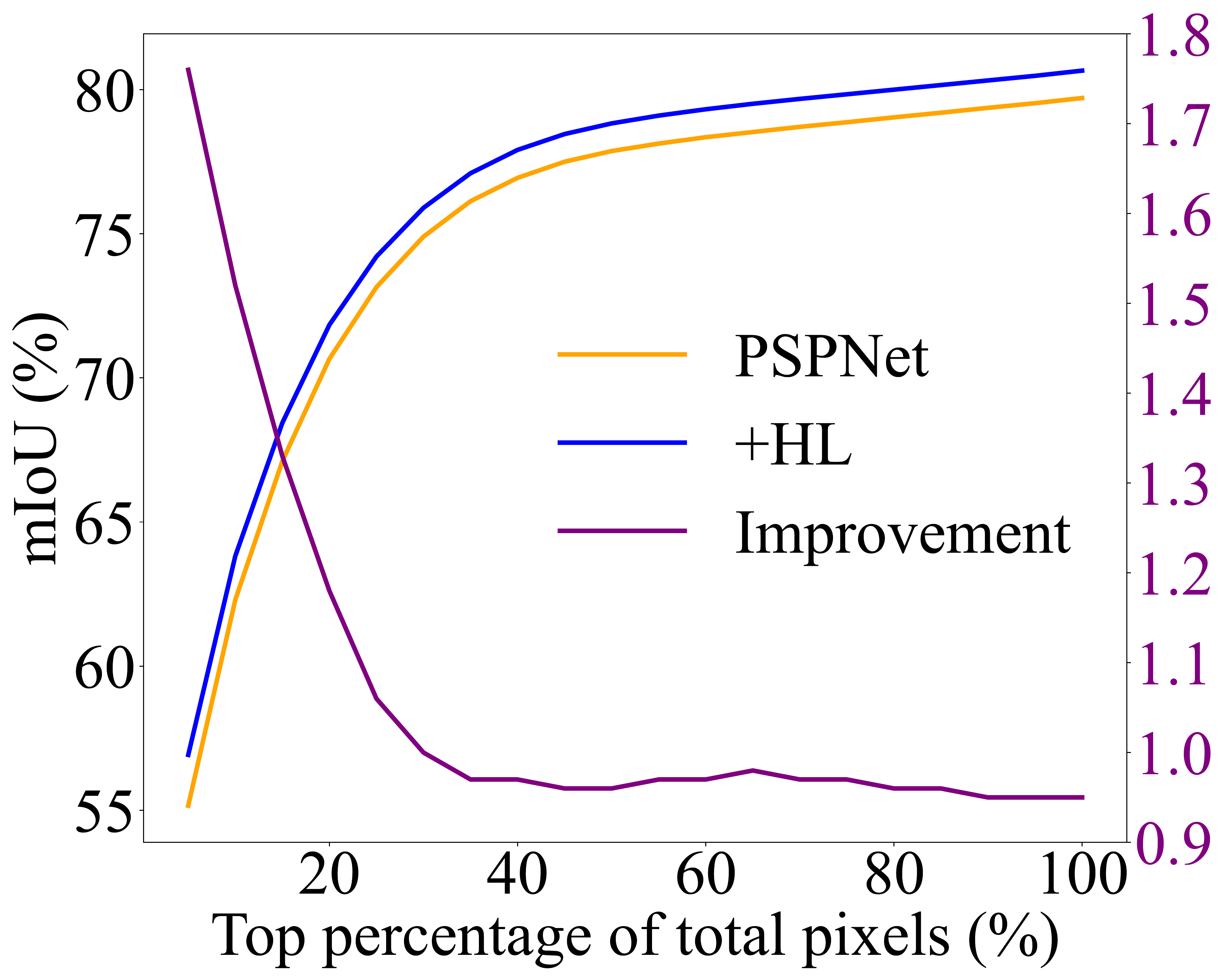}
	\end{minipage}}
        \hspace{1mm}
    \subfloat[ADE20K]{
        \begin{minipage}[b]{0.47\linewidth}
            \includegraphics[width=\linewidth]{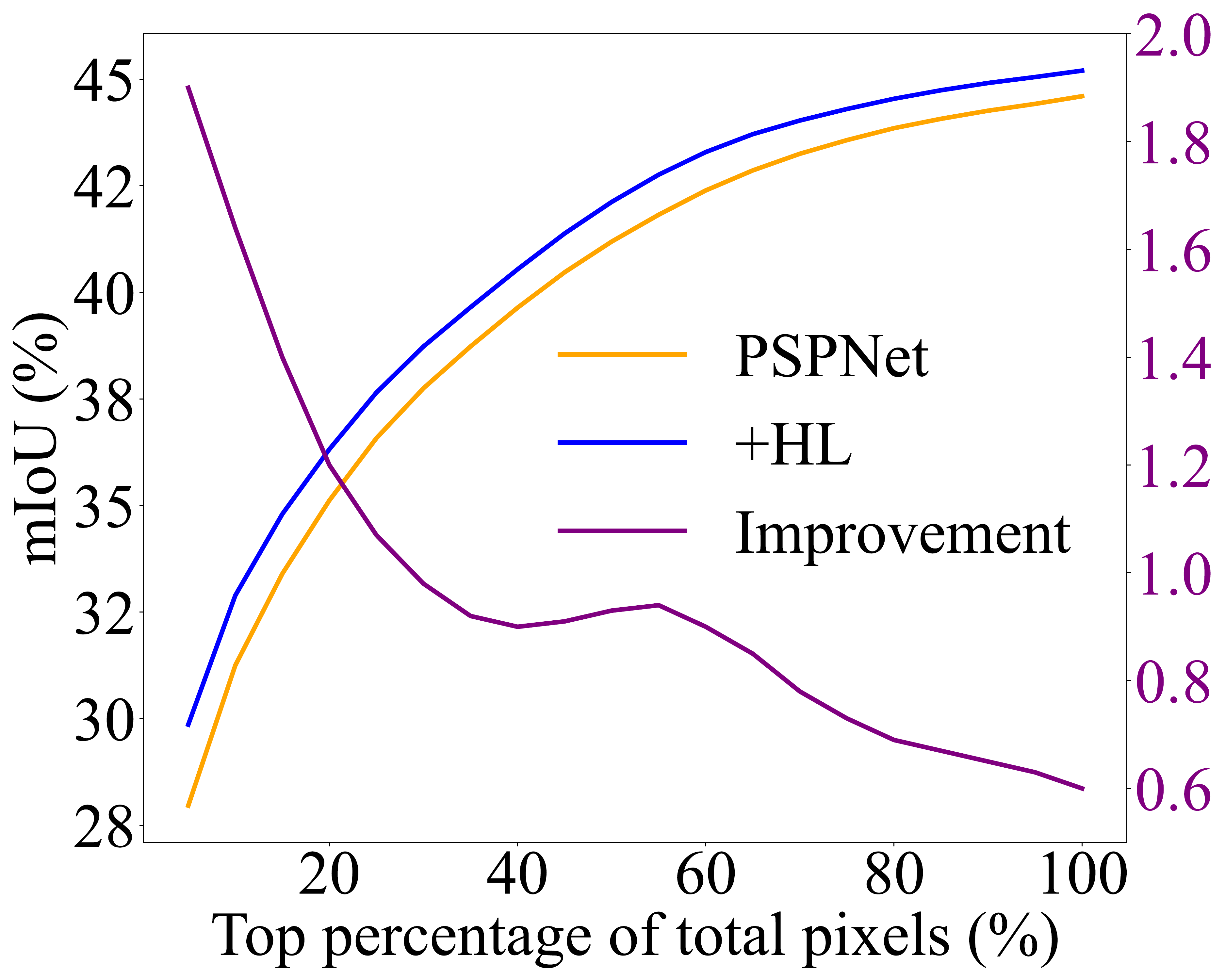}
        \end{minipage}}
    \caption{Quantitative evaluation of applying the proposed pixel
      hardness learning method to the PSPNet~\cite{zhao2017pyramid} on
      different pixel subsets with decreasing hardness on the
      validation set of Cityscapes and ADE20K.}
  \label{fig:analysis}
\end{figure}

\begin{table*}[tp]
  \centering
  \scriptsize
  \caption{Quantitative benchmarks of applying the proposed method to
    some segmentation methods with different backbone networks (from
    lightweight to cumbersome) on the validation set of
    Cityscapes. $*$ (\resp $\dag$) stands for training with
    $512\times1024$ (\resp $1024\times1024$) crop size, while testing
    on the whole image. SI: sidewalk; BU: building; TL: traffic light;
    TS: traffic sign; VE: vegetation; TE: terrain; PE: person; MO:
    motorcycle; BI: bicycle.}
  \setlength{\tabcolsep}{0.5mm}
  \resizebox{\textwidth}{!}
  {
  \begin{tabular}{l|l|cccccccccccccccccccc}
    \toprule
        Backbone &Method &Road &SI &BU  &Wall &Fence   &Pole    &TL     &TS  &VE    &TE    &Sky    &PE   &Rider    &Car &Truck    &Bus    &Train   &MO &BI &mIoU \\

    \midrule
    \multirow{2}*{ResNet-18~\cite{he2016deep}} &UPerNet~\cite{xiao2018unified} &97.72 &82.36 &91.80  &55.57 &56.47 &60.35 &63.80 &74.00 &91.82 &63.79 &94.32 &78.84 &57.06  &94.20 &64.94 &83.77 &71.22 &57.25 &74.21 &74.39 \\
     &+HL &97.68 &82.40 &92.04 &54.06 &59.00 &62.51 &66.61 &75.16 &92.01 &63.61 &94.31 &79.94 &60.21 &94.44 &69.53 &85.86 &77.33 &63.29 &75.87 &76.10 (\textcolor{violet}{+1.71})\\
    \midrule
    HRNetV2P-W18~\cite{wang2021deep} &OCRNet~\cite{YuanCW20} &98.07 &84.97 &92.05 &57.73 &59.55 &64.00 &66.71 &77.04 &92.15 &60.25 &94.45 &79.94 &57.28 &94.40 &81.20 &85.56 &74.31 &56.41 &74.00 &76.32 \\
     (Small)$^*$ &+HL &97.99 &84.81 &92.15 &53.18 &61.73 &66.26 &69.63 &77.10 &92.40 &62.50 &94.14 &81.62 &59.76 &94.80 &81.64 &87.66 &77.35 &61.76 &76.83 &77.54 (\textcolor{violet}{+1.22})\\
    \midrule
    \multirow{2}*{HRNetV2P-W18$^*$~\cite{wang2021deep}} &OCRNet~\cite{YuanCW20} &98.30 &86.04 &92.63 &59.50 &62.57 &66.90 &69.56 &78.39 &92.61 &64.10 &94.70 &82.06 &62.28 &95.14 &83.91 &88.49 &77.61 &60.38 &76.03 &78.48 \\
     &+HL &98.24 &86.09 &92.98 &59.10 &65.07 &69.02 &71.84 &79.29 &92.90 &64.59 &94.82 &83.57 &64.97 &95.26 &80.59 &88.61 &76.92 &63.11 &77.99 &79.21 (\textcolor{violet}{+0.73})\\
    \midrule
    \multirow{2}*{MobileNetV2$^*$~\cite{sandler2018mobilenetv2}} & PSPNet~\cite{zhao2017pyramid}  & 96.14 & 76.66 & 89.55 &38.73 & 50.14 & 59.31 & 56.94 & 72.93 & 90.39 & 45.96 & 92.37 & 77.65 & 52.53 & 92.77 &54.30  & 65.43 & 63.35 & 53.01 & 72.51 & 68.46 \\
     &+HL & 96.09 & 74.72 & 90.41 &48.54 & 48.68 & 60.76 & 61.82 & 73.70  & 90.88 & 49.31 & 92.65 & 78.68 & 54.04 & 93.11 &48.38 & 68.24 & 65.22 & 53.41 & 73.95 & 69.61 (\textcolor{violet}{+1.15}) \\
    \midrule
    \multirow{2}*{MobileNetV3$^*$~\cite{howard2019searching}} & LRASPP~\cite{howard2018inverted}  & 97.44 & 80.92 & 90.49 & 53.36 & 52.82 & 57.24 & 56.56 & 69.06 & 91.44 & 61.12 & 93.86 & 74.65 &49.02 & 92.63 & 59.04 & 75.34 & 56.50  & 50.99 & 70.87 & 70.18 \\
     &+HL & 97.45 & 81.37 & 90.69 & 51.98 & 54.54 & 60.23 &62.90  & 71.44 & 91.69 & 59.25 & 93.07 & 77.93 &54.73 & 92.98 & 60.43 & 77.35 & 56.72 & 55.58 & 73.84 & 71.80 (\textcolor{violet}{+1.62}) \\
    \midrule
    \multirow{2}*{BiSeNetV2$^{\dag}$~\cite{yu2021bisenet}} & BiSeNetV2~\cite{yu2021bisenet} &98.01 &83.45 &91.94 &57.69 &56.75 &60.68 &66.67 &76.17 &92.04 &62.01 &94.67 &79.42 &56.58 &94.29 &68.62 &73.61 &42.90 &56.31 &73.60 &72.92 \\
     &+HL & 97.92 &83.28 &91.81 &47.45 &56.98 &62.19 &68.82 &76.58 &92.13 &60.44 &94.54 &80.63 &59.41 &94.45 &78.17 &81.86 &71.26 &56.82 &75.46 &75.27 (\textcolor{violet}{+2.35}) \\
    \bottomrule
    \toprule
    \multirow{2}*{ResNet-50~\cite{he2016deep}} &UPerNet~\cite{xiao2018unified} &97.85 &83.52 &92.90 &61.76 &60.95 &64.28 &70.36 &78.66 &92.52 &65.79 &94.96 &81.78 &62.51 &95.02 &66.09 &87.97 &79.64 &66.32 &77.58 &77.92 \\
    & +HL &97.88 &83.84 &93.05 &56.46 &63.03 &66.68 &72.86 &80.37 &92.57 &64.63 &95.12 &83.13 &63.99 &95.39 &69.70 &87.97 &80.48 &68.62 &78.61 &78.65 (\textcolor{violet}{+0.73}) \\
    \midrule
    \multirow{2}*{ResNet-101~\cite{he2016deep}} &UPerNet~\cite{xiao2018unified} &98.12 &85.10 &92.95 &59.59 &63.82 &65.89 &71.63 &79.47 &92.64 &64.27 &95.00 &82.38 &62.40 &95.14 &72.98 &88.23 &81.65 &66.22 &78.07 &78.71 \\
     &+HL &98.11 &85.05 &93.26 &63.87 &64.41 &67.51 &73.68 &81.02 &92.87 &65.01 &95.06 &83.58 &65.72 &95.52 &75.71 &89.02 &83.03 &68.45 &79.60 &80.02 (\textcolor{violet}{+1.31}) \\
    \midrule
    \multirow{2}*{HRNetV2P-W48$^*$~\cite{wang2021deep}} &OCRNet~\cite{YuanCW20} &98.29 &85.86 &93.32 &58.99 &64.57 &69.55 &73.18 &81.28 &92.97 &66.43 &95.09 &83.94 &65.47 &95.70 &81.79 &92.16 &82.48 &69.41 &78.68 &80.48 \\
     &+HL &98.51 &87.75 &93.76 &64.65 &66.86 &71.94 &74.85 &82.14 &93.19 &66.68 &95.26 &84.85 &67.65 &96.01 &87.41 &92.42 &86.22 &70.56 &80.45 &82.17 (\textcolor{violet}{+1.69}) \\
   \midrule
   \multirow{2}*{ResNeXt-101~\cite{xie2017aggregated}} & UPerNet~\cite{xiao2018unified} & 98.18 & 85.42 & 93.28 & 63.12 & 65.80 & 67.68 & 72.95 & 80.90 & 92.78 & 66.06 & 95.13 & 83.86 & 66.35 & 95.64 & 76.03 & 89.42 & 83.92 & 67.66 & 79.17 & 80.18 \\
    & +HL & 98.18 & 85.44 & 93.57 & 63.59 & 66.21 & 69.49 & 74.58 & 82.18 & 92.85 & 66.06 & 95.24 & 84.56 & 67.86 & 95.70 & 74.23 & 89.14 & 83.27 & 70.82 & 80.17 & 80.69 (\textcolor{violet}{+0.51}) \\
    \midrule
   \multirow{2}*{ResNeSt-101~\cite{zhang2022resnest}} &UPerNet~\cite{xiao2018unified} & 98.09 & 84.75 & 93.14 & 64.03 & 63.35 & 66.13 & 71.19 & 80.21 & 92.73 & 65.08 & 95.08 & 82.80 & 63.68 & 95.38 & 73.43 & 86.90 & 82.01 & 67.16 & 78.04 & 79.11 \\
    & +HL & 98.10 & 84.67 & 93.40 & 64.03 & 66.02 & 68.01 & 73.99 & 81.03 & 92.81 & 63.66 & 95.12 & 83.82 & 66.08 & 95.82 & 77.05 & 90.81 & 84.45 & 70.62 & 79.70 & 80.48 (\textcolor{violet}{+1.37}) \\
    \midrule
    \multirow{2}*{ConvNeXt-Base~\cite{liu2022convnet}} & UPerNet~\cite{xiao2018unified} & 98.27 & 86.05 & 93.24 & 62.70  & 65.12 & 66.63 & 71.91 & 80.66 & 92.88 & 66.66 & 95.10  & 83.41 & 65.74 & 95.69 & 87.06 & 90.60  & 84.54 & 70.93 & 79.65 & 80.89 \\
     &+HL & 98.29 & 86.33 & 93.43 & 62.85 & 65.67 & 68.06 & 73.61 & 81.39 & 93.05 & 66.87 & 95.14 & 84.12 & 66.74 & 95.91 & 85.18 & 91.33 & 84.94 & 72.27 & 80.14 & 81.33 (\textcolor{violet}{+0.44}) \\
    \bottomrule
  \end{tabular}
  }
  \label{tab:backbone_cityscapes}
\end{table*}

\medskip
\noindent\textbf{Effectiveness of HL map in indicating hard pixels: }
The learned HL map is expected to be able to indicate hard pixels. To
verify this, we evaluate the segmentation performance with respect to
pixels with decreasing of HL. As depicted in Fig.~\ref{fig:analysis},
for both the ResNet-101-based baseline PSPNet~\cite{zhao2017pyramid}
and the proposed method trained on the corresponding training set, the
segmentation performance increases with respect to the decrease of
learned HL on the validation set of Cityscapes and ADE20K,
respectively. This implies that the learned HL map is effective in
indicating hard pixels. Besides, since the validation images are not
seen during training, this also implies that the proposed HL learning
generalizes well to unseen images. It is noteworthy that the proposed
hardness-aware semantic segmentation based on the learned HL map
improves the segmentation performance more on hard pixels (see
Fig.~\ref{fig:analysis}), further demonstrating the effectiveness of
the learned HL map in indicting hard pixels.

\medskip
\noindent\textbf{Generalization ability of HL map to unseen images: }
The proposed HL learning branch is normally only involved in the
training phase for semantic segmentation. Since the learned HL map is
stable and meaningfully related to image structure. The learned HL map
also generalizes well to unseen images. As shown in the middle column
of the bottom two rows of Fig.~\ref{fig:hardness} and the bottom two rows of
Fig.~\ref{fig:hardness_ade}, the proposed HL learning produces HL maps
with similar structures on unseen validation images. Besides, as shown
in the bottom two rows of Fig.~\ref{fig:hardness}, the model trained on
the training and validation set of Cityscapes produces very similar HL
maps with the model trained only on 
Cityscapes training set. This implies that the proposed HL learning generalizes
well to unseen images. Furthermore, using the model trained on the
ADE20K training set also yields similar HL maps (on Cityscapes
validation images) as the model trained on Cityscapes training set,
further demonstrating the good generalization ability of the proposed
HL learning.

To quantitatively evaluate the generalization ability of the proposed
HL learning, we compute the average of structural similarity
(SSIM)~\cite{wang2004image} between HL maps given by different models
on the Cityscapes validation set. Specifically, when applying the
proposed method to the PSPNet~\cite{zhao2017pyramid} with ResNet-101
backbone, we achieve 0.95 SSIM between the model trained on both
training and validation set of Cityscapes and the model trained only
on Cityscapes training set. This suggests that the proposed HL
learning generalizes well to unseen images of the same
dataset. Moreover, we get 0.84 SSIM between the model trained on
ADE20K training set and the model trained on Cityscapes training set,
further demonstrating the good generalization ability of the proposed
HL learning across datasets.

\begin{figure*}[tp]
  \centering
  \includegraphics[width=\linewidth]{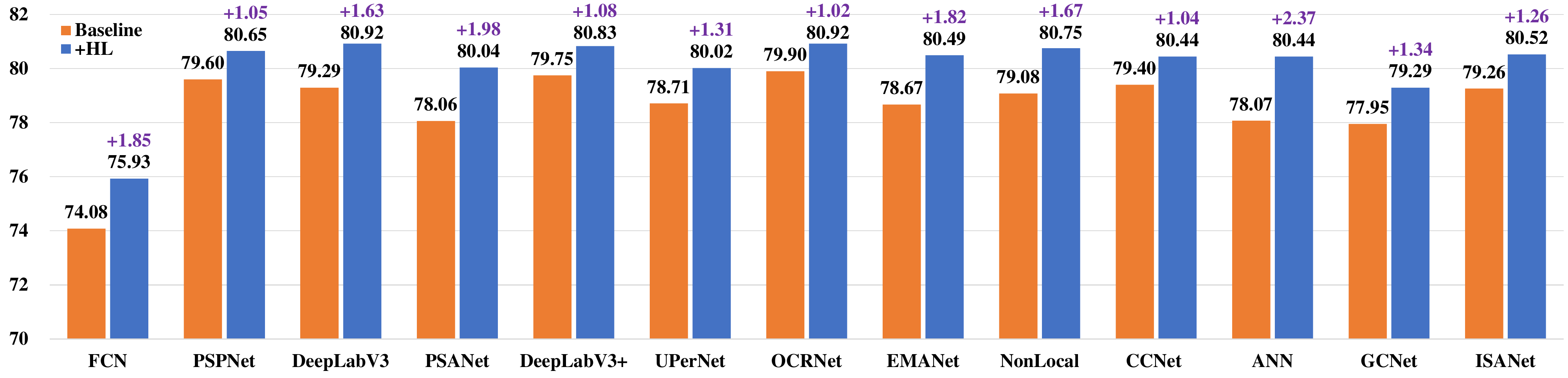}
  \caption{Quantitative benchmark of applying the proposed pixel
    hardness learning method on different popular segmentation methods
    with ResNet-101 as the backbone network. The proposed method
    achieves consistent or significant improvements with no extra cost
    during inference. }
  \label{fig:cityscape-res101}
\end{figure*}

\subsection{Experimental results}
\label{subsec:results}

\begin{figure*}[!tbp]
\centering
    \subfloat[Image]{
    	\begin{minipage}[b]{0.244\linewidth}
            \includegraphics[width=1\linewidth]{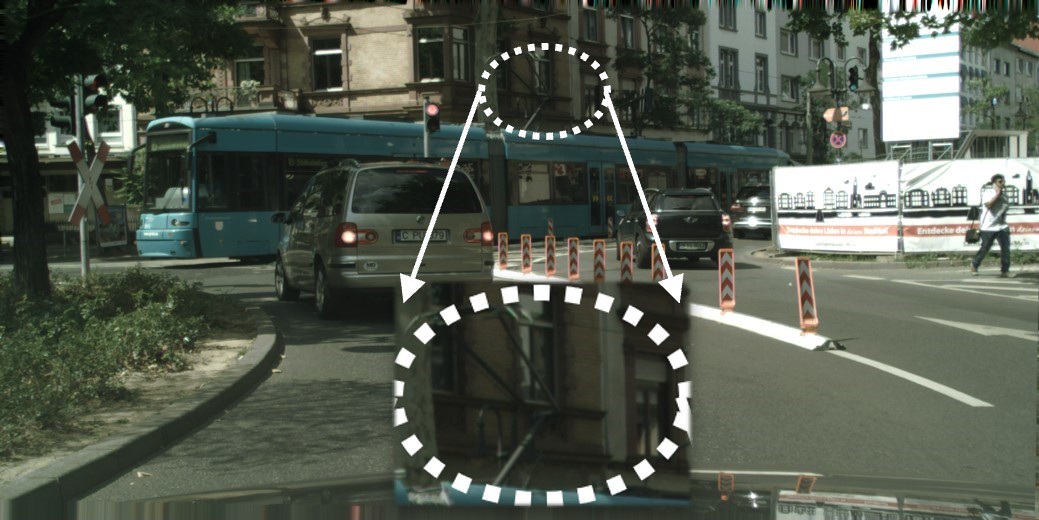}\vspace{0.75mm}
    		\includegraphics[width=1\linewidth]{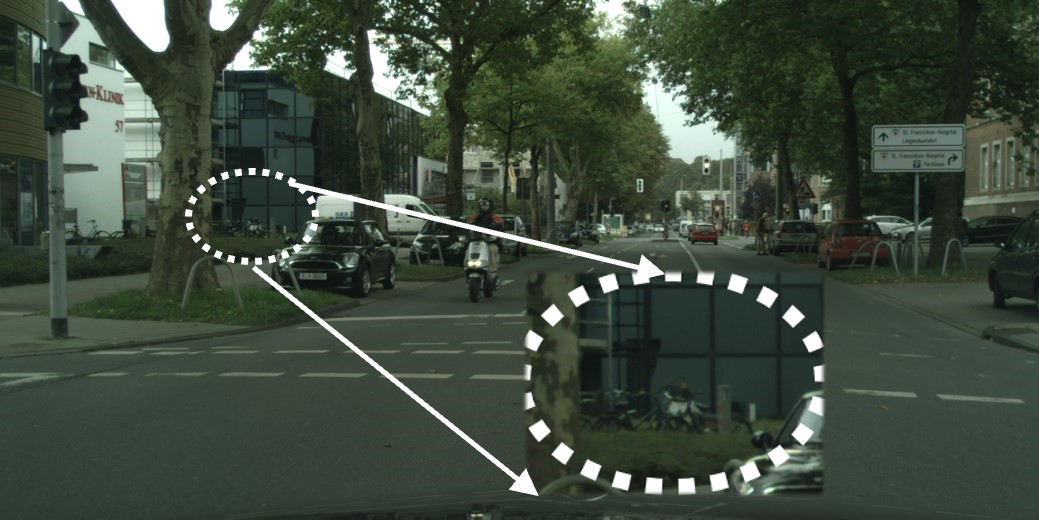}\vspace{1.5mm}
            \includegraphics[width=1\linewidth]{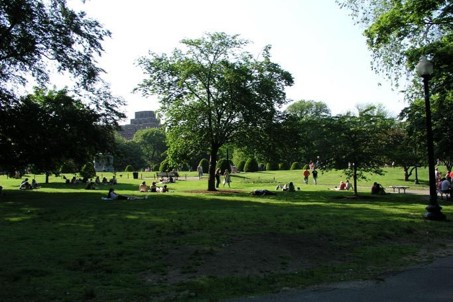}\vspace{0.75mm}
    		\includegraphics[width=1\linewidth]{figures/experiments/effectiveness/vis/ade/1.jpg}\vspace{1.5mm}
            \includegraphics[width=1\linewidth]{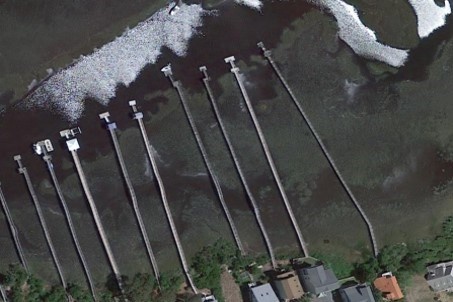}\vspace{0.75mm}
    		\includegraphics[width=1\linewidth]{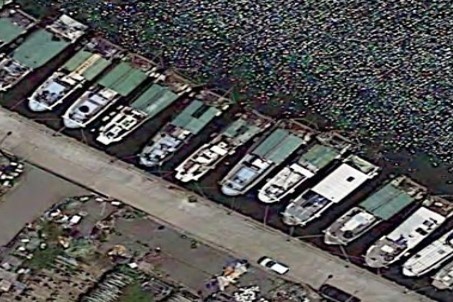}\vspace{1.5mm}
    		\includegraphics[width=1\linewidth]{figures/experiments/effectiveness/vis/text/2.jpg}\vspace{0.75mm}
    		\includegraphics[width=1\linewidth]{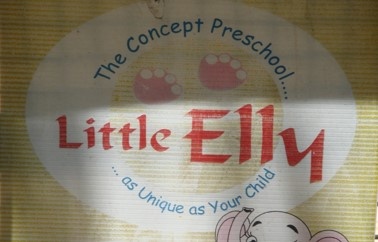}\vspace{0.75mm}
    	\end{minipage}
    }
    \hspace{-2mm}
    \subfloat[Ground truth]{
    	\begin{minipage}[b]{0.244\linewidth}
            \includegraphics[width=1\linewidth]{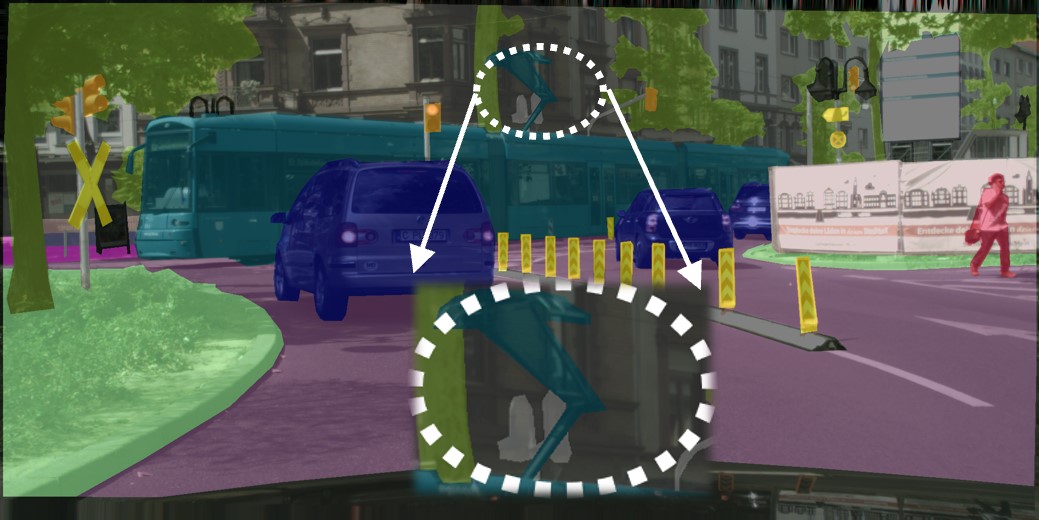}\vspace{0.75mm}
    		\includegraphics[width=1\linewidth]{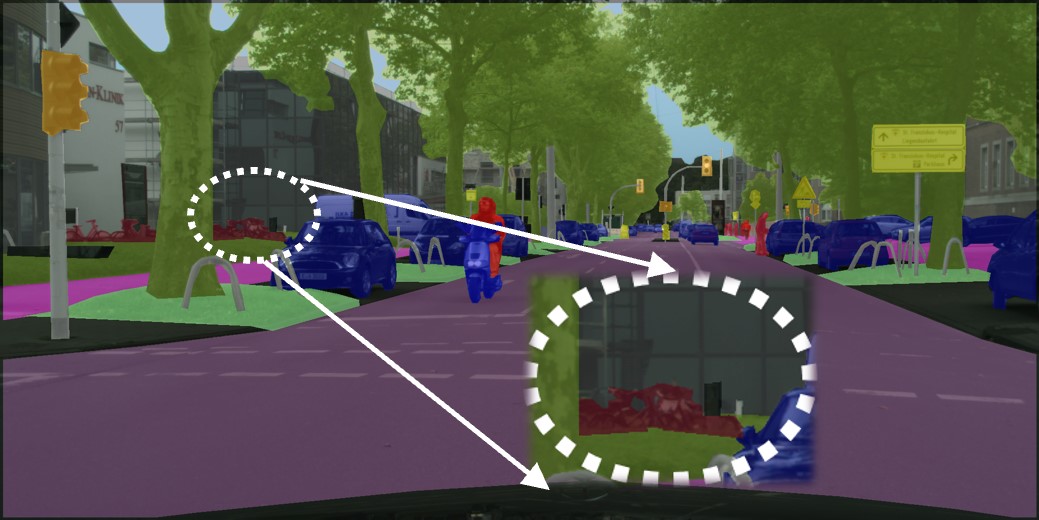}\vspace{1.5mm}
            \includegraphics[width=1\linewidth]{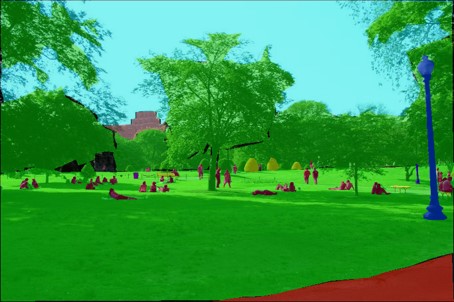}\vspace{0.75mm}
    		\includegraphics[width=1\linewidth]{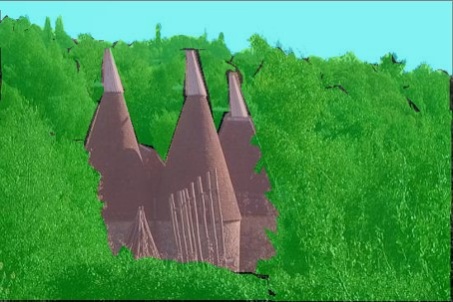}\vspace{1.5mm}
            \includegraphics[width=1\linewidth]{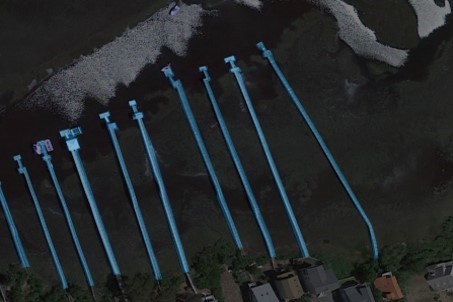}\vspace{0.75mm}
    		\includegraphics[width=1\linewidth]{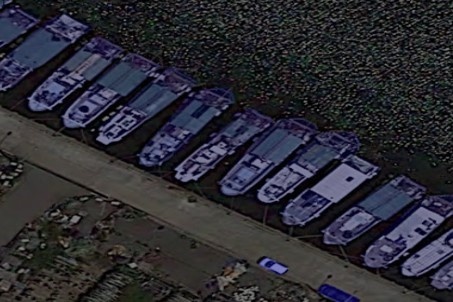}\vspace{1.5mm}
    		\includegraphics[width=1\linewidth]{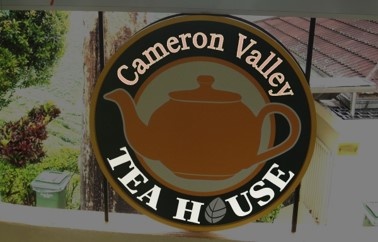}\vspace{0.75mm}
    		\includegraphics[width=1\linewidth]{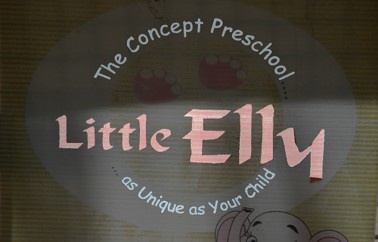}\vspace{0.75mm}
    	\end{minipage}
    }
    \hspace{-2mm}
    \subfloat[Baseline]{
    	\begin{minipage}[b]{0.244\linewidth}
            \includegraphics[width=1\linewidth]{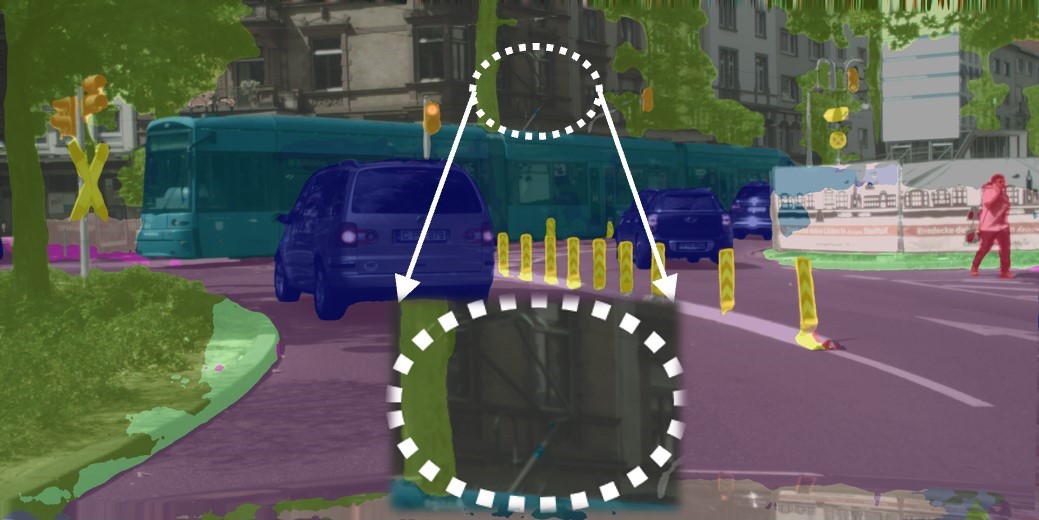}\vspace{0.75mm}
    		\includegraphics[width=1\linewidth]{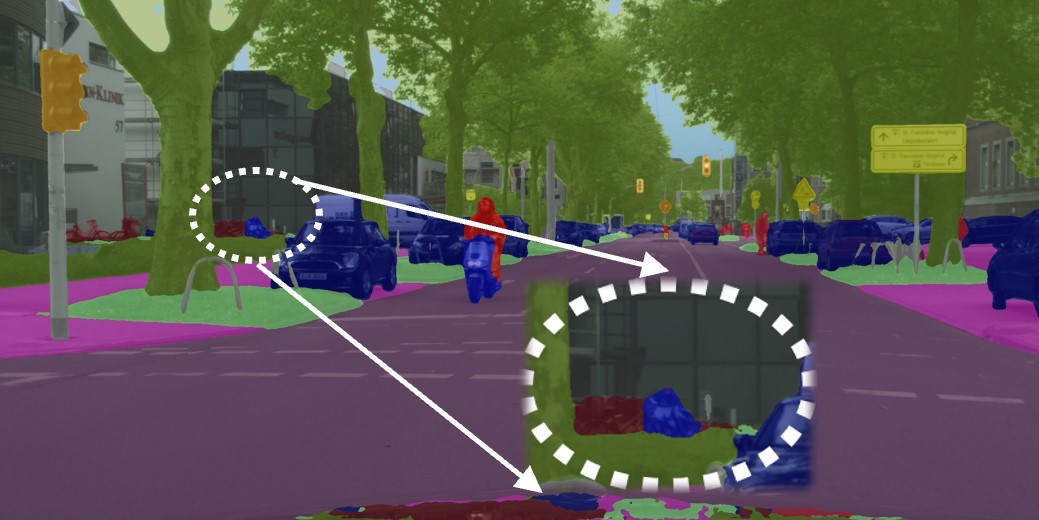}\vspace{1.5mm}
            \includegraphics[width=1\linewidth]{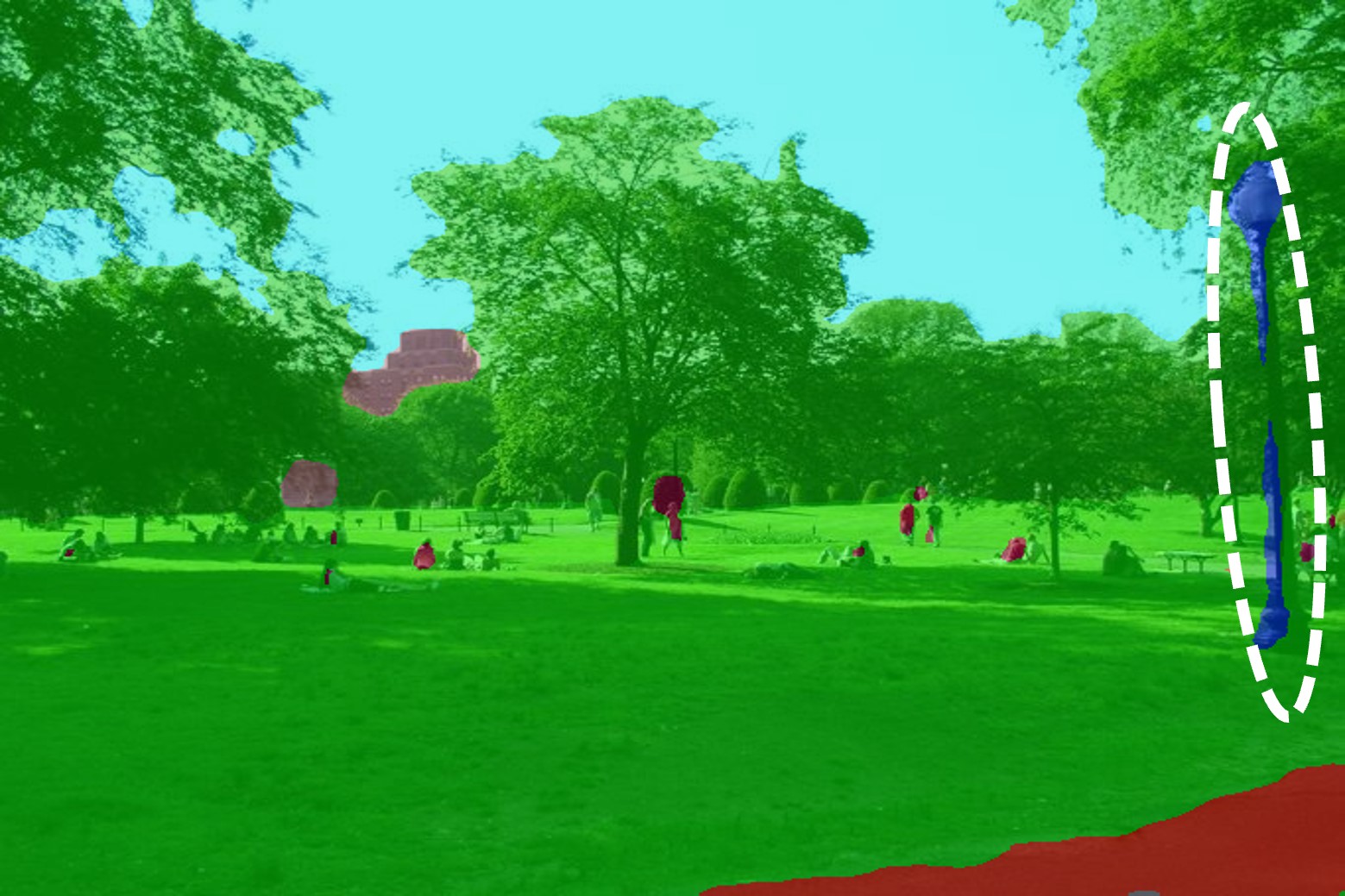}\vspace{0.75mm}
    		\includegraphics[width=1\linewidth]{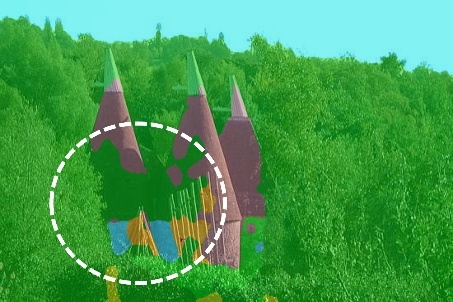}\vspace{1.5mm}
            \includegraphics[width=1\linewidth]{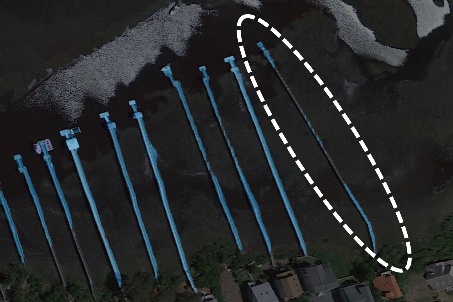}\vspace{0.75mm}
    		\includegraphics[width=1\linewidth]{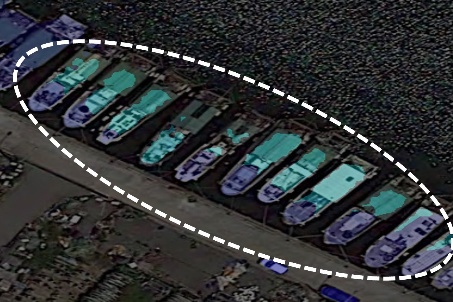}\vspace{1.5mm}
    		\includegraphics[width=1\linewidth]{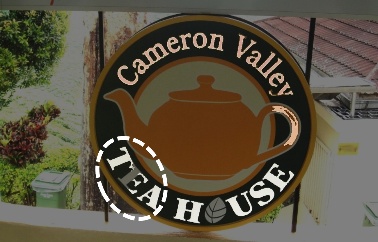}\vspace{0.75mm}
    		\includegraphics[width=1\linewidth]{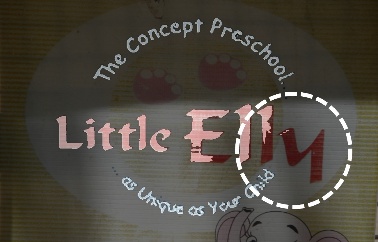}\vspace{0.75mm}
    	\end{minipage}
    }
    \hspace{-2mm}
    \subfloat[Baseline + HL (Ours)]{
    	\begin{minipage}[b]{0.244\linewidth}
            \includegraphics[width=1\linewidth]{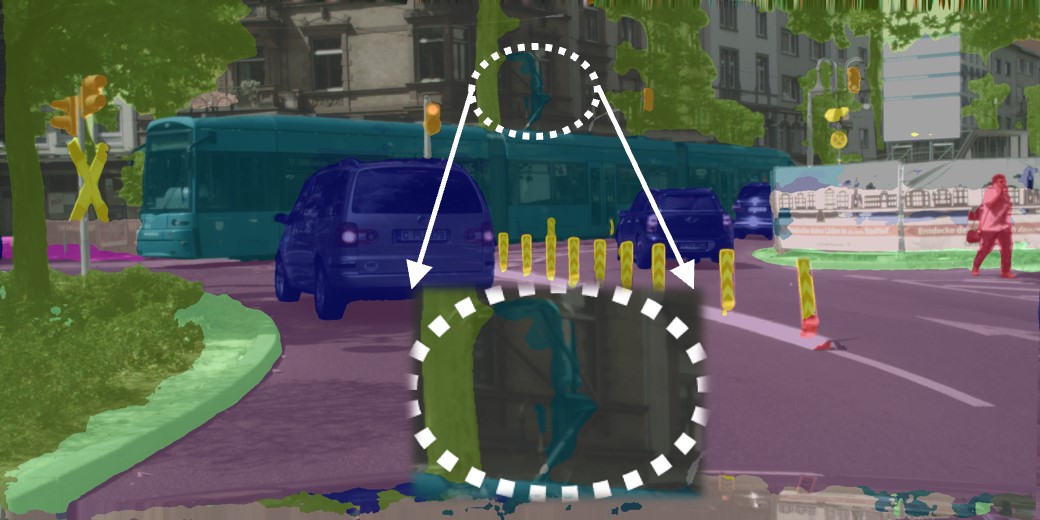}\vspace{0.75mm}
    		\includegraphics[width=1\linewidth]{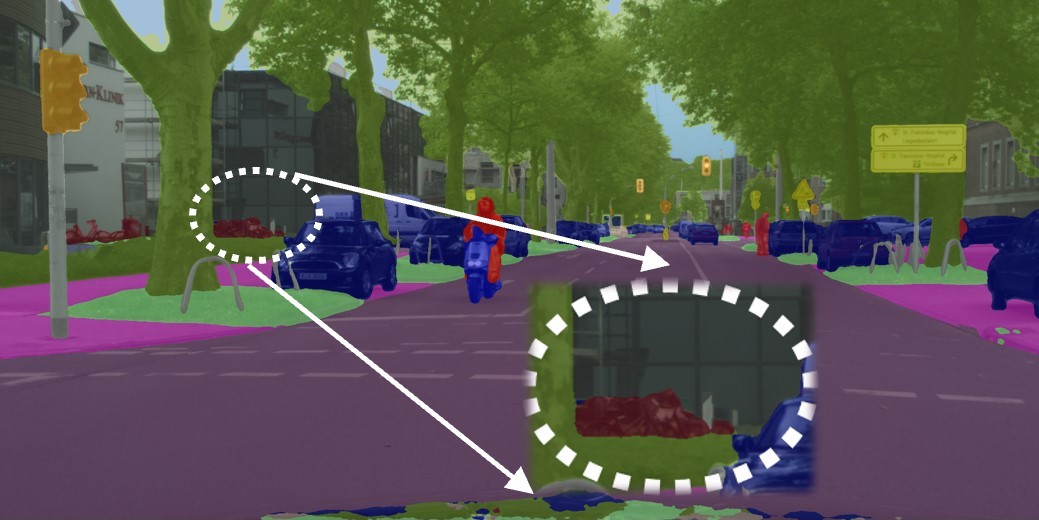}\vspace{1.5mm}
            \includegraphics[width=1\linewidth]{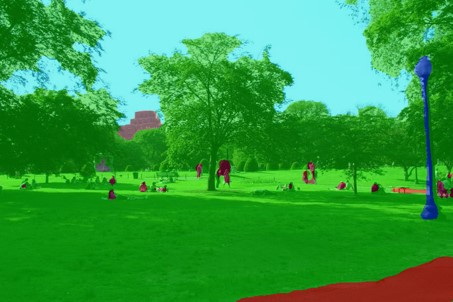}\vspace{0.75mm}
    		\includegraphics[width=1\linewidth]{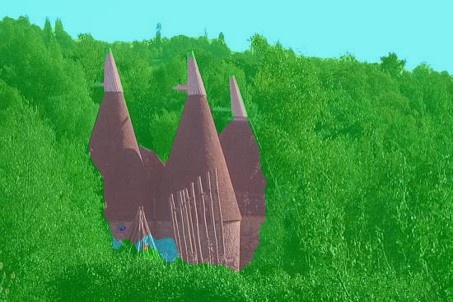}\vspace{1.5mm}
            \includegraphics[width=1\linewidth]{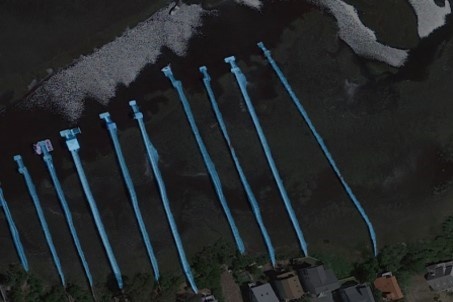}\vspace{0.75mm}
    		\includegraphics[width=1\linewidth]{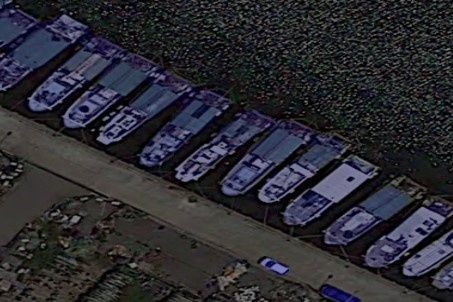}\vspace{1.5mm}
    		\includegraphics[width=1\linewidth]{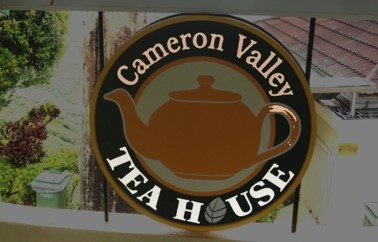}\vspace{0.75mm}
    		\includegraphics[width=1\linewidth]{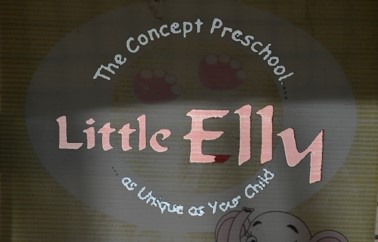}\vspace{0.75mm}
    	\end{minipage}
    }
    \caption{Qualitative illustration of some segmentation results on Cityscapes, ADE20K, iSAID, and Total-Text dataset (from top to bottom).}
  \label{fig:vis}
\end{figure*}

We conduct extensive experiments on four public datasets for semantic
segmentation. Firstly, we apply the proposed method to different
backbone networks and different baseline segmentation methods on
Cityscapes. This is followed by extensive experiments on ADE20K
dataset using some popular baseline segmentation methods and backbone
networks. We then evaluate the proposed method on iSAID and Total-Text
dataset for aerial and text image segmentation, respectively. Some
qualitative segmentation results are illustrated in
Fig.~\ref{fig:vis}. Applying the proposed HL learning approach to the
baseline method is able to accurately segment the image, including the
thin parts and small objects (within white ellipse of
Fig.~\ref{fig:vis}).  The corresponding quantitative benchmarks on
these datasets are given in the following.

\medskip
\noindent
\textbf{Experimental results on Cityscapes:} We first conduct two
types of extensive experiments on Cityscapes dataset.  The
quantitative results of using different backbones and different
segmentation architectures are detailed in the following.

\vspace{1.5mm}
\noindent\textit{Results of using different backbones:} There are many
different backbone networks (ranging from lightweight
ResNet-18~\cite{he2016deep} and MobileNet
family~\cite{sandler2018mobilenetv2, howard2019searching} to
cumbersome ResNet-101~\cite{he2016deep} and
ConvNeXt~\cite{liu2022convnet}) for semantic segmentation. We mainly
adopt widely used UPerNet~\cite{xiao2018unified} as the basic
segmentation architecture for most backbones. For some specifically
designed backbone networks, the adopted segmentation architectures in
the corresponding original papers are used. The quantitative benchmark
of applying the proposed HL learning method to different segmentation
backbones is depicted in Tab.~\ref{tab:backbone_cityscapes}. The
proposed method consistently or significantly outperforms the
corresponding baseline methods. Specifically, the proposed method
achieves in average 1.46\% mIoU improvement on lightweight backbones,
and 1.01\% mIoU improvement on cumbersome backbones. It is noteworthy that the
proposed method consistently/significantly improves the baseline
methods on categories of small-size objects (\eg, traffic light,
person, and rider) and with thin parts (\eg, pole, motocycle, and
bicycle). These quantitative results show that the proposed method is
effective in using various backbone networks for semantic segmentation.

\begin{table}[tbp!]
  \centering
  \caption{Quantitative evaluation of different popular methods on the
    validation set of ADE20K. $^*$ denotes backbone pre-trained on
    the ImageNet-22k other than the ImageNet-1k, and training with
    $640\times640$ crop size.}
  \setlength{\tabcolsep}{1.5mm}
  \renewcommand{\arraystretch}{1.2}
  {
  \begin{tabular*}{\linewidth}{l|lcc}
    \toprule
    Method  &Backbone &Baseline &+HL \\
    \midrule
    PSPNet~\cite{zhao2017pyramid}       &ResNet-50~\cite{he2016deep}    &42.04 &42.70 (\textcolor{violet}{+0.66}) \\
    PSPNet~\cite{zhao2017pyramid}       &ResNet-101~\cite{he2016deep}   &44.60 &45.20 (\textcolor{violet}{+0.60}) \\
    DeepLabV3+~\cite{chen2018encoder}   &ResNet-101~\cite{he2016deep}   &45.12 &45.43 (\textcolor{violet}{+0.31})\\
    UPerNet~\cite{xiao2018unified}      &ResNet-101~\cite{he2016deep}   &44.02 &44.55 (\textcolor{violet}{+0.53})\\
    CCNet~\cite{huang2020ccnet}         &ResNet-101~\cite{he2016deep}   &44.26 &44.61 (\textcolor{violet}{+0.35})\\
    OCRNet~\cite{YuanCW20}              &HRNetV2P-W48~\cite{wang2021deep}&43.20 &44.16 (\textcolor{violet}{+0.96})\\
    \midrule
    UPerNet~\cite{xiao2018unified}      &Swin-tiny~\cite{liu2021swin}   &43.50 &43.98 (\textcolor{violet}{+0.48})\\
    UPerNet~\cite{xiao2018unified}      &Swin-Base~\cite{liu2021swin}   &50.25 &50.95 (\textcolor{violet}{+0.70})\\
    UPerNet*~\cite{xiao2018unified}      &Swin-Base$^*$~\cite{liu2021swin} &51.29 &52.52 (\textcolor{violet}{+1.23})\\
    UPerNet*~\cite{xiao2018unified}      &ConvNeXt-Base$^*$~\cite{liu2022convnet} &52.17 &52.96 (\textcolor{violet}{+0.79})\\
    \bottomrule
  \end{tabular*}
  }
  \label{bac_arch_ade}
\end{table}

\vspace{1.5mm}
\noindent\textit{Results of using different segmentation
  architectures:} Since the pioneering FCN~\cite{long2015fully} for
semantic segmentation, numerous fully convolutional semantic
segmentation methods have been proposed. To further assess the
effectiveness of the proposed method, we apply the proposed method on
various popular segmentation architectures using the same
ResNet-101~\cite{he2016deep} backbone. As shown in
Fig.~\ref{fig:cityscape-res101}, the proposed method also achieves
consistent improvements ranging from 1.02\% to 2.37\% mIoU (1.49\%
mIoU on average) over all baseline methods. These results demonstrate
that the proposed method can be applied to most segmentation methods
and brings consistent improvements.

\medskip
\noindent
\textbf{Experimental results on ADE20K:} Since the experiment on ADE20K
requires 160k iterations rather than 40k iterations for Cityscapes, we
mainly conduct experiments by applying the proposed method to some
classical segmentation methods using popular backbone networks
(\eg, ResNet-50/ResNet-101~\cite{he2016deep},
HRNet~\cite{wang2021deep}, Swin Transformer~\cite{liu2021swin}, and
ConvNeXt~\cite{liu2022convnet}). As depicted in
Tab.~\ref{bac_arch_ade}, the proposed method achieves consistent
improvements (0.66\% mIoU on average) over all baseline methods. In
particular, applying the proposed method to
UPerNet~\cite{xiao2018unified} with Swin
Transformer~\cite{liu2021swin} as the backbone network achieves 1.23\%
mIoU improvement over the baseline method with relatively superior
performance. The experimental results on ADE20K also demonstrate the
effectiveness of the proposed method on various segmentation methods
with different backbone networks.

\begin{table}[tbp!]
  \centering
  \caption{Quantitative results of some methods on the validation set of iSAID.}
  \renewcommand{\arraystretch}{1.2}
  {
  \begin{tabular*}{\linewidth}{l|lcc}
    \toprule
    Method  &Backbone &Baseline &+HL \\
    \midrule

    FCN~\cite{long2015fully}   &HRNet-W18 (small)~\cite{wang2021deep}  &62.80 &63.74 (\textcolor{violet}{+0.94})\\

    FCN~\cite{long2015fully}   &HRNet-W18~\cite{wang2021deep}  &65.75 &66.60 (\textcolor{violet}{+0.85})\\
    \midrule

    PSPNet~\cite{zhao2017pyramid}   &ResNet-18~\cite{he2016deep}  &60.76 &62.93 (\textcolor{violet}{+2.17})\\

    PSPNet~\cite{zhao2017pyramid}   &ResNet-50~\cite{he2016deep}  &65.65 &66.50 (\textcolor{violet}{+0.85})\\
    \bottomrule
  \end{tabular*}
  }
  \label{bac_arch_isaid}
\end{table}

\medskip
\noindent
\textbf{Experiments on iSAID:} Aerial images are usually acquired in
top-down view, and are thus quite different from natural images. There
are numerous small objects in the large-scale aerial images. Thus, it
is quite challenging to segment high-resolution aerial images. We
conduct experiments on the iSAID\cite{waqas2019isaid} for
aerial image segmentation. Because of high-resolution images,
following the common practice for this dataset, we simply conduct
experiments using two popular segmentation methods with some compact
backbones. The quantitative comparison with the baseline methods is
depicted in Tab.~\ref{bac_arch_isaid}. The proposed method
achieves consistent improvements (1.20\% mIoU on average)
in segmenting aerial images, demonstrating the generality of the
proposed method in segmenting various types of images.

\begin{table}[tbp!]
  \centering
  \caption{Quantitative results of some methods on the test set of
    Total-Text. The IoU score for the foreground text is depicted.}
  \setlength{\tabcolsep}{1.6mm}
  \renewcommand{\arraystretch}{1.2}
  {
    \label{text}
  \begin{tabular*}{\linewidth}{l|lcc}
    \toprule
    Method  &Backbone &Baseline &+HL \\
    \midrule

    FCN~\cite{long2015fully}   &HRNet-W18 (small)~\cite{wang2021deep}  &66.62 &68.56 (\textcolor{violet}{+1.94})\\

    FCN~\cite{long2015fully}   &HRNet-W18~\cite{wang2021deep}  &69.32 &70.58 (\textcolor{violet}{+1.26})\\
    \midrule

    PSPNet~\cite{zhao2017pyramid}   &ResNet-18~\cite{he2016deep}  &49.43 &50.53 (\textcolor{violet}{+1.10})\\

    PSPNet~\cite{zhao2017pyramid}   &ResNet-50~\cite{he2016deep}  &52.91 &53.93 (\textcolor{violet}{+1.02})\\
    \bottomrule
  \end{tabular*}
  }
\end{table}

\medskip
\noindent
\textbf{Experiments on Total-Text:} Finally, we evaluate the proposed
method in segmenting scene texts on Total-Text~\cite{CK2019}.
Following the common practice, we report the IoU score of the
foreground text (fgIoU) to quantitatively benchmark different
methods. We adopt the same baseline segmentation methods as
experiments on iSAID for aerial image segmentation. The quantitative
evaluation is depicted in Tab.~\ref{text}, where we observe similar
consistent performance improvements (1.33 fgIoU on average) for scene
text segmentation. This further demonstrates the generality and
effectiveness of the proposed method for semantic segmentation.

\begin{table}[tp]
  \centering
  \caption{Quantitative results of using different pixel weight
    strategies in PSPNet~\cite{zhao2017pyramid} on Cityscapes validation
    set, using different backbones.}
  \setlength{\tabcolsep}{2.5mm}
  \renewcommand{\arraystretch}{1.2}
  {
    \label{dif_method}
  \begin{tabular*}{\linewidth}{lccc}
    \toprule
    \multirow{2}*{Method} &\multicolumn{3}{c}{PSPNet~\cite{zhao2017pyramid} with different backbone networks} \\
    \cmidrule{2-4}      &ResNet-18 &ResNet-50 &ResNet-101\\
    \midrule
    Cross-Entropy  &74.30 &78.68 &79.60\\
    Balanced CE  &74.21 &78.18 &79.16\\
    OHEM~\cite{shrivastava2016training}            &74.23 &78.99 &79.64\\
    Focal loss~\cite{lin2017focal}          &73.35 &78.43 &78.95\\
   HL (Ours) &\textbf{75.60} &\textbf{79.58} &\textbf{80.65}\\
    \bottomrule
  \end{tabular*}
  }
\end{table}

\begin{table}[tp]
  \centering
  \caption{Quantitative results of setting different values for $c$
    involved in Eq.~\eqref{eq:hardnesslevel} on Cityscapes validation set,
    using different backbones for PSPNet~\cite{zhao2017pyramid}.. }
  \setlength{\tabcolsep}{2mm}
  \renewcommand{\arraystretch}{1.2}
  {
    \label{tab:aba_constant}
  \begin{tabular*}{\linewidth}{lccc}
    \toprule
    \multirow{2}*{Hyper-parameter $c$}  &\multicolumn{3}{c}{PSPNet~\cite{zhao2017pyramid} with different backbone networks}\\
    \cmidrule{2-4} &ResNet-18   &ResNet-50  &ResNet-101\\
    \midrule
    Baseline  &74.30  &78.68  &79.60 \\
    $c = 0.01$  &75.93 (\textcolor{violet}{+1.63}) &79.39 (\textcolor{violet}{+0.71}) &79.89 (\textcolor{violet}{+0.29}) \\
    $c = 0.1$  &75.60 (\textcolor{violet}{+1.30})  &79.58 (\textcolor{violet}{+0.90})  &80.65 (\textcolor{violet}{+1.05}) \\
    $c = 0.5$  &74.82 (\textcolor{violet}{+0.52})  &79.04 (\textcolor{violet}{+0.36})  &80.41 (\textcolor{violet}{+0.81}) \\
    \bottomrule
  \end{tabular*}
  }
\end{table}

\subsection{Ablation study}
\label{sec:ablation}
We conduct three types of ablation studies on Cityscapes: the
effectiveness of using different pixel weights for pixel-wise
cross-entropy segmentation loss, and the influence of the two
hyper-parameters involved in Eq.~\eqref{eq:hardnesslevel} and
Eq.~\eqref{eq:lossfinal}.

\medskip
\noindent\textbf{Ablation study on different pixel weights:} We mainly
compare the proposed hardness level learning method with three types
of pixel weights: 1) the balanced cross-entropy (Balanced CE)
loss that weigh the classical cross-entropy loss based on the number
of pixels in each category; 2) Cross-entropy loss with
online hard example mining~\cite{shrivastava2016training} (OHEM CE).
Following the default setting in mmsegmentation~\cite{mmseg2020}, we
set the probability threshold to 0.7 and keep at least 100,000 pixels
for training; 3) Focal loss~\cite{lin2017focal}, for which we set the
parameter $\gamma$ to default value 2.

As depicted in Tab.~\ref{dif_method}, for the
PSPNet~\cite{zhao2017pyramid} with different backbone networks, using
the other three alternative pixel weights in cross-entropy
segmentation loss does not perform well, and even sometimes performs
worse than the baseline segmentation loss. This is not
surprising and explains why they are not widely used in semantic
segmentation. Indeed, as described in Sec.~\ref{subsec:overview}, the
OHEM CE and focal loss are based on the current loss values. This may
make the model over-fit to very few pixels which are ambiguous to
distinguish, yielding degenerated segmentation results.  On the other
hand, the proposed hardness level learning method achieves consistent
performance improvement, showing the effectiveness of the proposed
method.

\medskip
\noindent \textbf{Ablation study on hyper-parameter $c$ in
  Eq.~\eqref{eq:hardnesslevel}:} This hyper-parameter $c$ controls the
maximum relative ratio between different pixels, helping to avoid
overwhelming hardness level for a few pixels. A lower value for $c$
implies focusing only on harder pixels for segmentation. We conduct
an ablation study on $c$ by setting it to 0.01, 0.1, and 0.5 for the
PSPNet~\cite{zhao2017pyramid} with different backbone networks. As
depicted in~\ref{tab:aba_constant}, the segmentation performance
varies for different values of $c$, and is consistently better than
the baseline method. This demonstrates that the proposed method is
rather effective. Following the results of using the ResNet-101
backbone, we simply set $c$ to 0.01 for all experiments. It is
noteworthy that using a smaller value for $c$ results in better
segmentation performance improvement for the segmentation model with
lightweight ResNet-18, which is less powerful than the cumbersome
backbone in fitting the training samples. There are more hard pixels
for the lightweight segmentation model. Therefore, using a smaller value
of $c$ for the lightweight segmentation backbone focuses more on
harder pixels, yielding better results. In practice, we could adjust
the value of $c$ for different segmentation backbones to get better
results.

\begin{table}[tp]
  \centering
  \caption{Quantitative results of setting different values for
    $\alpha$ involved in Eq.~\eqref{eq:lossfinal} on Cityscapes
    validation set, using different backbones for
    PSPNet~\cite{zhao2017pyramid}.}
  \setlength{\tabcolsep}{2mm}
  \renewcommand{\arraystretch}{1.2}
  {
  \label{tab:hyper-parameter_alpha}
  \begin{tabular*}{\linewidth}{lccc}
    \toprule
    \multirow{2}*{Hyper-parameter $\alpha$}  &\multicolumn{3}{c}{PSPNet~\cite{zhao2017pyramid} with different backbone networks}\\
    \cmidrule{2-4} &ResNet-18   &ResNet-50  &ResNet-101\\
    \midrule
    Baseline &74.30 &78.68 &79.60 \\
    $\alpha = 0.001$ &74.37 (\textcolor{violet}{+0.07}) &78.69 (\textcolor{violet}{+0.01}) &79.86 (\textcolor{violet}{+0.26})\\
    $\alpha = 0.01$ &75.60 (\textcolor{violet}{+1.30}) &79.58 (\textcolor{violet}{+0.90}) &80.65 (\textcolor{violet}{+1.05})\\
    $\alpha = 0.1$ &75.94 (\textcolor{violet}{+1.64}) &79.62 (\textcolor{violet}{+0.94})  &80.44 (\textcolor{violet}{+0.84})\\
    $\alpha = 1$ &75.82 (\textcolor{violet}{+1.52}) &79.88 (\textcolor{violet}{+1.20}) &80.16 (\textcolor{violet}{+0.56})\\
    \bottomrule
  \end{tabular*}
  }
\end{table}

\medskip
\noindent \textbf{Ablation study on hyper-parameter $\alpha$ in
  Eq.~\eqref{eq:lossfinal}:} This hyper-parameter controls the
increasing rate of competing hardness level. We set $\alpha$ to 0.001,
0.01, 0.1, and 1 for PSPNet~\cite{zhao2017pyramid} with different
backbone networks. As listed in Tab.~\ref{tab:hyper-parameter_alpha},
all settings of $\alpha$ improve the baseline method. For $\alpha
\geq 0.01$, the proposed method achieves in general noticeable
performance improvement for all backbone networks. Following the
segmentation result with ResNet-101, we set $\alpha$ to 0.01 for all
experiments.

\begin{table}[tb]
  \centering
  \caption{Analysis of extra training cost (in terms of GPU memory
    usage) for the proposed method based on experiments on the
    Cityscapes dataset.}
  \setlength{\tabcolsep}{6mm}
  \renewcommand{\arraystretch}{1.2}
  {
  \label{tab:cost}
  \begin{tabular*}{\linewidth}{l|ll}
    \toprule
    Method &Backbone  &Increasing\\
    \midrule
    NonLocal~\cite{wang2018non} &\multirow{5}*{ResNet-101~\cite{wang2018non}} &0.68\% \\
    PSPNet~\cite{zhao2017pyramid} & &1.40\% \\
    DeepLabV3+~\cite{chen2018encoder} &  &4.26\% \\
    UPerNet~\cite{xiao2018unified} & &12.11\% \\
    \midrule
    LRASPP~\cite{howard2018inverted} &MobileNetV3~\cite{howard2019searching} &0.01\% \\
    BiSeNetV2~\cite{yu2021bisenet} &BiSeNetV2~\cite{yu2021bisenet} &0.20\% \\
    PSPNet~\cite{zhao2017pyramid} &MobileNetV2~\cite{sandler2018mobilenetv2} &0.59\% \\
    \bottomrule
  \end{tabular*}
  }
\end{table}

\subsection{Complexity analysis}
\label{subsec:complexity}
The proposed hardness level learning for semantic segmentation is only
involved during training. The performance improvements in
Sec.~\ref{subsec:results} is achieved with no
extra cost during the test phase. The proposed method only requires some extra
cost in training. We give in Tab.~\ref{tab:cost} the extra cost in
terms of relatively increased GPU memory usage for some segmentation
methods on Cityscapes. As depicted in Tab.~\ref{tab:cost}, the
proposed method requires in general ignorable extra cost during
training phase, which makes it easily applicable for most semantic
segmentation methods.

\subsection{Application to other segmentation tasks}
\label{subsec:application}
Though the proposed method is mainly developed for fully-supervised
semantic segmentation task, it can be easily applied to other
segmentation tasks such as semi-supervised and domain-generalized
semantic segmentation.

\begin{table}[tp]
  \centering
  \caption{Quantitative results of applying the proposed method to
    U$^2$PL~\cite{wang2022semi} for semi-supervised semantic
    segmentation on Cityscapes validation set. ``SupOnly" denotes the
    unlabeled images are not used for training. $^{\dagger}$ stands for
    results from the original paper~\cite{wang2022semi}.}
  \setlength{\tabcolsep}{0.5mm}
  \renewcommand{\arraystretch}{1.2}
  {
  \label{tab:u2pl}
  \begin{tabular*}{\linewidth}{l|cccc}
    \toprule
    \multirow{2}*{Method} &\multicolumn{4}{c}{Portion (number) of training images} \\
    \cmidrule{2-5} &1/16 (186)  &1/8 (372) &1/4 (744) &1/2 (1488)\\
    \midrule
    SupOnly$^{\dagger}$ &65.74 &72.53 &74.43 &77.83 \\
    \midrule
    U$^{2}$PL (w/CutMix)$^{\dagger}$ &70.30 &74.37 &76.47 &79.05 \\
    \midrule
    U$^{2}$PL (w/CutMix) &71.11 &75.24 &75.86 &78.43  \\
    +HL &72.62 \scriptsize(\textcolor{violet}{+1.51}) &76.04 \scriptsize(\textcolor{violet}{+0.80}) &76.55 \scriptsize(\textcolor{violet}{+0.69}) &79.63 \scriptsize(\textcolor{violet}{+1.20})   \\
    \bottomrule
  \end{tabular*}
  }
\end{table}

\medskip
\noindent
\textbf{Semi-supervised semantic segmentation:} Semi-supervised
semantic segmentation aims to perform segmentation by making use of
few labeled images and many unlabeled images. We conduct experiments
of semi-supervised semantic segmentation on Cityscapes using
state-of-the-art U$^2$PL~\cite{wang2022semi} as the baseline model.
All the images in the training set of Cityscapes are used during
training, but only a portion of the annotations of these training
images are used. We compare the results under classical 1/16, 1/8,
1/4, and 1/2 partition protocols in semi-supervised semantic
segmentation. As shown in Tab.~\ref{tab:u2pl}, the proposed method
also achieves consistent and noticeable performance improvements
(1.05\% mIoU) in semi-supervised semantic segmentation on
Cityscapes. Note that we reproduce the results for U$^2$PL using its
official implementation for a fair comparison.  This shows that the
proposed method is also effective for semi-supervised semantic
segmentation.

\begin{table}[tp]
  \centering
  \caption{Quantitative benchmark of different domain-generalized
    semantic segmentation methods.  ``Extra data'' denotes using extra
    real-world data during training. $\dag$ denotes results from the original paper. }
  \setlength{\tabcolsep}{4mm}
  \renewcommand{\arraystretch}{1.2}
  {
  \label{tab:dg}
  \begin{tabular*}{\linewidth}{l|cc}
    \toprule
    Method &Extra data &GTAV $\Rightarrow$ Cityscapes\\
    \midrule
    ISW~\cite{choi2021robustnet}$\dag$ \tiny{(CVPR2021)} &\ding{55} &37.09 \\
    FSDR~\cite{huang2021fsdr}$\dag$ \tiny{(CVPR2021)} &\ding{51} & 44.80 \\
    WildNet~\cite{lee2022wildnet}$\dag$ \tiny{(CVPR2022)} &\ding{55} &45.79 \\
    SHADE~\cite{zhao2022shade}$\dag$ \tiny{(ECCV2022)} &\ding{55} &46.66 \\
    \midrule
    DeepLabV3+~\cite{chen2018encoder} &\ding{55} &36.56  \\
    +HL &\ding{55} &43.06 \scriptsize(\textcolor{violet}{+6.50}) \\

    \bottomrule
  \end{tabular*}
  }
\end{table}

\medskip
\noindent
\textbf{Domain-generalized semantic segmentation:} To improve the
segmentation performance for unseen domains, domain-generalized semantic
segmentation (DG-Seg) has recently attracted much attention. We
conduct a simple experiment to verify the effectiveness of the
proposed method in DG-Seg task. Specifically, following the other
DG-Seg methods~\cite{choi2021robustnet, lee2022wildnet,
  zhao2022shade}, we adopt DeepLabV3+~\cite{chen2018encoder} with
ResNet-101 as the baseline model. We train both the baseline
and the proposed method on the synthetic dataset
GTAV~\cite{richter2016playing}, and report segmentation results on the
validation set of Cityscapes. As shown in Tab.~\ref{tab:dg}, without
bells and whistles, the proposed method achieves a 6.50\% mIoU
improvement over the baseline model in generalizing from synthetic
GTAV to Cityscapes. The proposed method without any DG strategies
achieves 43.06 mIoU, which is on par with some recent methods
specifically designed for the DG-Seg task.

\section{Conclusion}
\label{sec:conclusion}
In this paper,we propose a novel pixel hardness learning
method for semantic segmentation. Differently from existing hard pixel mining based on current loss
values, the proposed pixel hardness learning makes use of global and
historical loss values. This results in a stable and meaningful hardness
level map related to inherent image structure, which generalizes well
and helps in segmenting difficult areas. Applying the proposed pixel
hardness learning method to many popular semantic segmentation methods
achieves consistent/significant improvements on natural image
segmentation, aerial and text image segmentation. Besides, the
proposed method also improves the state-of-the-art semi-supervised
semantic segmentation method, and demonstrates good generalization
ability across domains. Note that the proposed method requires no
extra cost during inference, and only slightly increases the training
cost. In the future, we would like to explore the idea of pixel
hardness learning for more applications, such as object detection and
other dense prediction tasks.

\ifCLASSOPTIONcaptionsoff
  \newpage
\fi

\renewcommand{\baselinestretch}{.95}

\bibliographystyle{IEEEtran}

\bibliography{reference}



\end{document}